\documentclass[lettersize,journal]{IEEEtran}
\usepackage{amsmath,amsfonts}
\usepackage{algorithm}
\usepackage{algorithmic}
\usepackage{array}
\usepackage[caption=false,font=normalsize,labelfont=sf,textfont=sf]{subfig}
\usepackage{float}
\usepackage{textcomp}
\usepackage{stfloats}
\usepackage{url}
\usepackage{verbatim}
\usepackage{graphicx}
\usepackage{cite}
\usepackage{ragged2e}
\DeclareMathOperator*{\argmax}{arg\,max}
\usepackage[colorinlistoftodos]{todonotes}

\usepackage[normalem]{ulem}
\begin{document}

\title{Deep Reinforcement Learning with Explicit Context Representation}

\author{Francisco Munguia-Galeano,
Ah-Hwee Tan,~\IEEEmembership{Senior Member,~IEEE,}

Ze Ji,~\IEEEmembership{Member,~IEEE}

\thanks{Manuscript received April 19, 2021; revised August 16, 2021. This work was partly supported by Consejo Nacional de Humanidades Ciencias y Tecnologías (CONAHCyT) and Cardiff University. (Corresponding author: jiz1@cardiff.ac.uk)} \thanks{Francisco Munguia-Galeano and Ze Ji are with the School of Engineering, Cardiff University, Cardiff, CF24 3AA, United Kingdom (e-mail: \{MunguiaGaleanoF, jiz1\}@cardiff.ac.uk)}
\thanks{A.-H. Tan is with the School of Computing and Information Systems, Singapore Management University (email: ahtan@smu.edu.sg).
}}
\markboth{Journal of \LaTeX\ Class Files,~Vol.~14, No.~8, August~2021}%
{Francisco Munguia-Galeano \MakeLowercase{\textit{et al.}}: Deep Reinforcement Learning with Explicit Context Representation}

\maketitle

\begin{abstract}
Reinforcement learning (RL) has shown an outstanding capability for solving complex computational problems. However, most RL algorithms lack an explicit method that would allow learning from contextual information. Humans use context to identify patterns and relations among elements in the environment, along with how to avoid making wrong actions. On the other hand, what may seem like an obviously wrong decision from a human perspective could take hundreds of steps for an RL agent to learn to avoid. This paper proposes a framework for discrete environments called Iota explicit context representation (IECR). The framework involves representing each state using contextual key frames (CKFs), which can then be used to extract a function that represents the affordances of the state; in addition, two loss functions are introduced with respect to the affordances of the state. The novelty of the IECR framework lies in its capacity to extract contextual information from the environment and learn from the CKFs' representation. We validate the framework by developing four new algorithms that learn using context: Iota deep Q-network (IDQN), Iota double deep Q-network (IDDQN), Iota dueling deep Q-network (IDuDQN), and Iota dueling double deep Q-network (IDDDQN). Furthermore, we evaluate the framework and the new algorithms in five discrete environments. We show that all the algorithms, which use contextual information, converge in around 40,000 training steps of the neural networks, significantly outperforming their state-of-the-art equivalents. 
\end{abstract}

\begin{IEEEkeywords}
Artificial Intelligence, Deep Reinforcement Learning, Machine Learning, Neural Networks, Q-learning, Reinforcement Learning, Affordances.
\end{IEEEkeywords}

\section{Introduction}
\label{sec:introduction}

\begin{figure}[ht]
    \centering
            \includegraphics[width=3.5in]{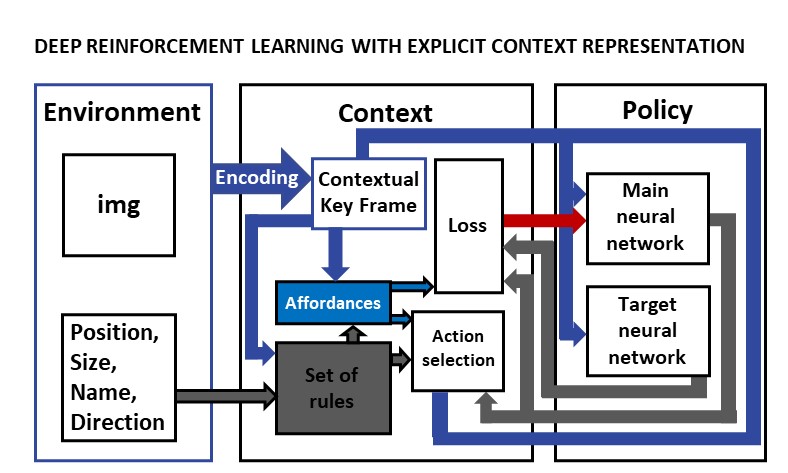}%
    \caption{Deep reinforcement learning with explicit context representation.}
    \label{fig:framework_1}
\end{figure}

During the past few years, reinforcement learning (RL), a sub-field of Machine learning (ML), has demonstrated successful capability in learning policies to control decision-making in sequential environments~\cite{schaul2015prioritized, hanna2021importance, wang2022deep}. One of the most famous RL algorithms is Q-learning (QL), which is an off-policy RL approach that updates the action selection for a given state using Bellman optimal equations that are based on the temporal difference (TD) principle. TD combines the Monte Carlo method and dynamic programming~\cite{watkins1992q}. In RL, the quality of data sampling significantly impacts the learning progress of the agent. Off-line RL learning approaches have shown that high-quality data improves RL performance~\cite{yu2021conservative, yang2021hierarchical, ren2018self}. Typically, RL algorithms utilize an $\epsilon$-greedy policy~\cite{mnih2015human}, which involves initially selecting actions randomly and then biasing the preference towards actions chosen by the agent.

Additionally, the quality of data sampling relies on the effectiveness of the exploration process. In this context, it is crucial to determine when and how to conduct data sampling~\cite{osband2019deep, bellemare2016unifying, pan2018multisource}. However, the main issue with state-of-the-art approaches is that undesirable data can become trivial due to the $\epsilon$-greedy policy. This is because the agent keeps selecting incorrect actions for multiple episodes until the $\epsilon$-greedy policy reaches a low probability, and the agent learns to avoid them. As a result, the training dataset grows exponentially with low-quality data, leading to a deceleration in the agent's learning process. This challenge is referred to as the efficiency problem in sampling~\cite{nair2018overcoming, blonde2019sample}.

A way to improve data sampling efficiency is by using contextual information. Context provides meaning to raw data, reduces ambiguity, and helps focus attention on a clear objective. Without context, a situation can be incomprehensible. In the scope of this paper, the term \textit{context} is defined as the set of conditions and circumstances associated with a given state in the environment. Many reinforcement learning environments have access to contextual data, but it requires significant time and effort for an RL agent to directly learn from this information.

In addition to the limited involvement of contextual information in RL, distinguishing between two states is challenging due to the stochastic nature of RL~\cite{ribeiro2002reinforcement, fujimoto2019off}. A critical knowledge gap lies in effectively leveraging available contextual information, such as affordances, objects, positions, shapes, and other physical characteristics, to enhance the sampling quality of the agent during environment exploration. Ideally, an agent should learn as rapidly as a human does~\cite{finn2017model, voss2020playing}. 

In this paper, we present the Iota Explicit Context Representation (IECR) framework, which aims to improve learning performance by incorporating contextual data (Fig.~\ref{fig:framework_1}). The IECR framework initially converts environment information into contextual key frames (CKFs), which consist of a matrix where each cell contains a token representing an element from the environment, along with its position, size, and direction. Our contributions can be summarized as follows:

\begin{itemize}
\item We introduce IECR, a framework that utilizes contextual data to enhance the learning process of RL agents.
\item We propose four new algorithms based on IECR: Iota Deep Q-network (IDQN), Iota Double Deep Q-network (IDDQN), Iota Dueling Deep Q-network (IDuDQN), and Iota Dueling Double Deep Q-network (IDDDQN).
\end{itemize}

We conduct two stages of experiments: (i) the first stage aims to demonstrate the impact of the CKFs representation on the learning performance of IDQN, and (ii) the second stage evaluates the performance of the proposed framework and state-of-the-art algorithms under the same conditions. Both stages utilize the implementations from stable-baselines~\cite{stable-baselines}. The benchmark problem domain for evaluating the algorithms consists of five discrete environments, and their specific characteristics are detailed in Section~\ref{sec:setup}.

The rest of the paper is organized as follows: Section~\ref{sec:related_work} introduces the relevant literature on DQN, DDQN, DuDQN, and DDDQN. Section~\ref{sec:background} reviews several relevant works and discusses the similarities and differences with our framework. In section~\ref{sec:iecr}, we formally introduce the framework and detail the new algorithms proposed. Section~\ref{sec:setup} introduces the characteristics of the environments and neural networks and the evaluation metrics employed. In section~\ref{sec:results}, we describe our results. Section~\ref{sec:discussion} presents a discussion. Finally, we propose future work and conclude the paper in section~\ref{sec:conclusion}.

\section{Related work}
\label{sec:related_work}

This section introduces popular deep RL approaches and relevant literature, summarized in three subsections: context-free, implicit context-based and explicit context-based methods.

\subsection{Context-free methods}

\begin{figure}[htbp]
    \centering
    \includegraphics[width=3.5in]{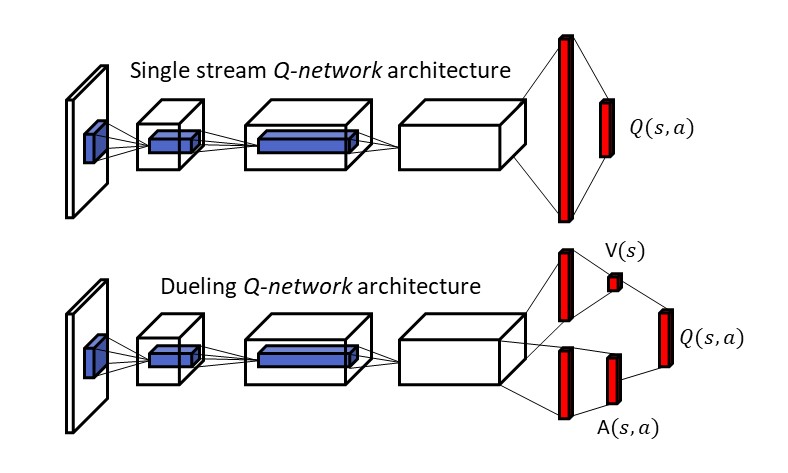}%
    \caption{Typical neural network architectures used in DQN's variants~\cite{wang2016dueling}.}
    \label{fig:sta_architectures}
\end{figure}

This subsection comprises methods in which no contextual information is given. Hence, the agent learns everything from scratch. Among these methods, Deep Q-network (DQN), the implementation of neural networks into QL~\cite{mnih2015human}, has several state-of-the-art variants such as Double deep Q-network (DDQN)~\cite{hasselt2010double}, Dueling deep Q-network (DuDQN)~\cite{wang2016dueling} and Double dueling deep Q-network (DDDQN)~\cite{wang2016dueling} (see Fig.~\ref{fig:sta_architectures}). Part of the success of those approaches can be attributed to their scalability.

Due to its versatility, DQN has been used for many applications and has proved its effectiveness in multiple fields~\cite{arulkumaran2017deep}. Successful applications of DQN and its variants include DQN for same-day deliveries~\cite{chen2022deep}, path planning for autonomous surface vehicles~\cite{luis2021multiagent,li2019deep,chai2022design,yang2018hierarchical}, and even in optimization of demand-side management systems for energy consumption~\cite{tai2019real}. One of the most well-known implementations of DQN is for playing Atari games~\cite{mnih2015human}, where DQN-trained policies outperformed humans in most of the games used for their experiments.

DQN and DDQN use two neural networks: a main neural network and a target neural network. The input for both networks is a given state, and the outputs are the Q-values. While the main neural network serves for training and making decisions, the target network stabilizes the agent's learning process (the target neural network must be frozen for a number of steps, usually 100). In RL, sampling adequate data from the training environment is particularly important as it improves the learning process. This is because the agent incrementally updates its parameters while producing its own training set~\cite{xu2021improving,deisenroth2011pilco}.

An approach to solve this problem is Deep Q-network from demonstrations (DQNfD)~\cite{hester2018deep}. DQNfD is an example of how the quality of the sampled data improves the agent's performance. As a way of illustration, DDNfD takes 1 million steps to achieve good scores, while DDQN takes 84 to 85 million steps to accomplish a similar performance. The approach creates a data set with human demonstrations to pre-train a neural network. After using the DQNfD algorithm, the agent outperforms DDQN in 41 out of 42 Atari games. However, these algorithms lack the human ability to recognize actions while observing video streams~\cite {liu2020fuzzy,liu2022human}, and this valuable contextual information is often omitted. This situation leaves the agent with the task of learning it from scratch. IECR combines the information of the CKFs and the set of rules to extract affordable sets of actions.

\subsection{Implicit context-based methods}

Several existing approaches use indirect contextual information and previous experience to enhance their learning capabilities. For example, Transfer learning (TL)~\cite{taylor2009transfer} is an approach that uses previous experiences from solving preliminary source tasks to learn a new policy and use it for a new target task. Therefore, TL uses fewer samples than if the policy has been trained from scratch. Given a target task, an RL transfer agent must perform three steps: select a correct source task, find the relation between source and target tasks, and transfer knowledge from source to target task. QL has been combined with TL in Transfer Reinforcement Learning under Unobserved Contextual Information~\cite{zhang2020transfer}, in which they propose to solve a contextual Markov decision process (CMDP)~\cite{hallak2015contextual,belogolovsky2021inverse} using TF. The authors provide casual bounds and a context-aware policy to assist the agent's learning process. On the contrary, IECR includes all the contextual information from the very beginning encoded in CKFs representing the states.   

Kabra et al.~\cite{kabra2021potent} demonstrates the potential of using a contextual exploration technique in real-time user recommendations. Hence, using previous experiences avoids training an agent from scratch. A similar approach is employed in~\cite{kabra2021cluster}, where  RL and contextual resources are used to make top-k recommendations. On the contrary, IECR starts from scratch without any previous experience by producing a context-aware selection of actions through the training process using only CKFs and an affordance function.  

Another approach to improve stochastic exploration performance is generative action selection through probability (GRASP)~\cite{xu2021improving}. GRASP uses a generator for exploration spaces by using a generative adversarial network (GAN). Under other conditions, context can be included indirectly in the agent's learning process. Methods such as proximal policy optimization (PPO)~\cite{schulman2017proximal} and asynchronous actor-critic (A2C)~\cite{mnih2016asynchronous} can be equipped with recurrent neural networks (RNN). RNNs use the previous output as part of the input. Hence, it is possible to store past information; in this way, RNNs feed the agent with indirect contextual data. In this work, PPO and A2C implementations from the stable-baselines are used to compare their performance with IDQN, IDDQN, IDuDQN, and IDDDQN.

\subsection{Explicit context-based methods}

In the literature, several methods exist that include explicit contextual information. For example, Benjamins et al.~\cite{benjamins2022contextualize,benjamins2021carl} introduced a framework designed to solve CMDPs and a benchmark library. The framework includes information such as gravity, target distance, actuator strength, and joint stiffness in the learning process. They demonstrated how the contextual data affect the performance learning of the agents. However, the affordances of the actions are not included in these studies and lack the codification of the state that IECR uses. An approach that involves encoding the state and the contextual features is presented by Sodhani et al.~\cite{sodhani2021multi}. The authors develop a method that relies on the capacity to relate tasks to external supplementary data to improve the agent's learning performance. Additionally, Injecting domain knowledge into the neural network~\cite{teng2014self} is another method that proves contextual data's positive influence on an agent's learning performance. However, all the methods mentioned above still fail to include important contextual information, such as the affordances of the actions.

In this context, affordance is the semantic link between the environment and the possibility of taking an action~\cite{gibson1977theory}. For example, a hammer affords to hit, while a pen affords to write. Affordances have been applied successfully to different fields~\cite{yamanobe2017brief,koppula2015anticipating}. An example of a successful implementation of affordances in an agent is in~\cite{chalmers2017learning}, which presents a method based on affordances to predict a human's next action such that a robot reacts accordingly. Moreover, the method not only reaches any environment where its predictions are valid but also accelerates RL in new scenarios. However, learning affordances in RL (usually done through a large number of interactions with the environment or learning from demonstration) still lacks an explicit method to add affordance rules based on context and the ability to use them to improve the learning process of the agent.

Particularly relevant to this paper is the approach called Training Agents with Interactive Reinforcement Learning and Contextual Affordances~\cite{cruz2016training}, which provides resources known as ``contextual affordances", aiming to use them as constraints of high-level actions depending on the elements present in the environment. Despite adding contextual affordances to assist the exploration task, the authors use the same loss function defined in~\cite{mnih2015human} rather than the context-reactive loss functions IECR uses. Another difference is that IECR provides CKFs that are part of the representation of the state and more information about the environment such that it is possible to feed the agent with CKFs rather than using a set of binary codes that represent a small set of elements in the environment that only assist in the exploration stage. To the best of the authors' knowledge, the tokenization of the state in the shape of CKFs to find the affordances and use both elements to train a neural network in discrete environments still need further investigation. 
    
\section{Preliminaries}
\label{sec:background}
A Markov decision process (MDP) is a 5-tuple $\langle S, A, R, T, \gamma \rangle$ where $S$ is a set of states, $A$ is a set of actions, $R (s, a)$ is a reward function, $T(s' |s, a)$ is a transition function equal to a probability distribution $P(s' |s, a)$, and $\gamma$ is a discount factor~\cite{sutton2018reinforcement}. An agent uses a policy $\pi$ to select an action $a \in A$ given a state $s \in S$; the agent then reaches state $s'$ according to $P(s' |s, a)$ and receives a reward $R(s, a)$. The agent's main objective is to maximize the expected discounted reward over the agent's trajectory, which implies finding an optimal policy $\pi^*$. The $Q$ function $Q^{\pi} (s, a)$ denotes the value of an action state pair that approximates the expected future reward that can be obtained from $(s, a)$ when a policy $\pi$ is given. The optimal value function $Q^{*} (s, a)$ can be obtained from the Bellman optimality equation~\cite{mnih2015human}:

\begin{equation}\label{eq:1}
Q^{*}(s,a) = \mathbb{E}_{s' \sim P}\Bigg[R(a,s)+\gamma  \max_{a^{'}}Q^{*}(s',a')\Bigg]
\end{equation}

The expectation can be removed and approximated using bootstrapping because the reward function $R(s, a)$  is the immediate reward $r$ obtained when performing an action $a$ in a given state. Therefore, the $Q$ function $Q_{\theta}(s,a)$ can be approximated with a neural network using $k$ transitions from a replay buffer $D^{replay}$; then, the optimal value function $Q^{*}(s,a)$ is:

\begin{equation}\label{eq:2}
Q^{*}(s,a) = r +\gamma\max_{a^{'}} Q^{*}(s',a')
\end{equation}

DQN uses two neural networks. The main neural network $\theta$ computes the present value of the given pair $(s, a)$. The target neural network $Q_{\theta'}(s,a)$  is frozen for $n$ training steps and is used to compute the next pair $(s',a')$; here, the loss function $J_{dn}(Q)$ is given by the following:

\begin{equation}\label{eq:3}
J_{dn}(Q) = r +\gamma\max_{a^{'}} Q^{*}(s',a') - Q_{\theta}(s,a)
\end{equation}   

DQN takes the maximum Q-value of the target neural network, while DDQN takes the index of the maximum Q-value from the main neural network and then takes the value of that index from the target neural network. In this way, DDQN solves the problem of overestimating the states that DQN presents in some environments. Since DQN sometimes learns unreasonable high values, the main goal of DDQN~\cite{hasselt2010double} is to reduce the overestimation of actions for a given state $s$. DDQN takes the maximum index of the main neural network and then the actual value of that index in the target neural network to calculate the loss $J_{ddn} (Q)$: 

\begin{equation}\label{eq:4}
J_{ddn}(Q)= r+\gamma Q_{\theta'}(s',\argmax_{a}Q_{\theta}(s',a)) - Q_{\theta}(s,a)
\end{equation} 

Dueling deep Q-network (DuDQN)~\cite{wang2016dueling} uses two output separate estimators: the advantage output stream $A_{\theta}(s, a)$ and the $V_{\theta}(s)$ output stream. DuDQN uses two streams in the neural network to estimate separately the advantage $A_{\theta}(s, a)$ that is given by the following:

\begin{equation}\label{eq:5}
A_{\theta}(s,a)=Q_{\theta}(s,a)-V_{\theta}(s),                         
\end{equation} 

\noindent where $A_{\theta}(s,a)$ is the advantage, and $V_{\theta}(s)$ is the value of the state $s$. Even though the actions are selected using the advantage stream channel, the loss function~\eqref{eq:3} for DuDQN is the same used in DQN. DDDQN is the implementation of the DDQN principle to DuDQN. DuDQN and DDDQN take actions based on the maximum value of the advantage stream output. Fig.~\ref{fig:sta_architectures} shows how the output streams from DQN and DDQN differ from the architectures of DuDQN and DDDQN.

\begin{figure*}[!t]
\centering
\includegraphics[width=7in]{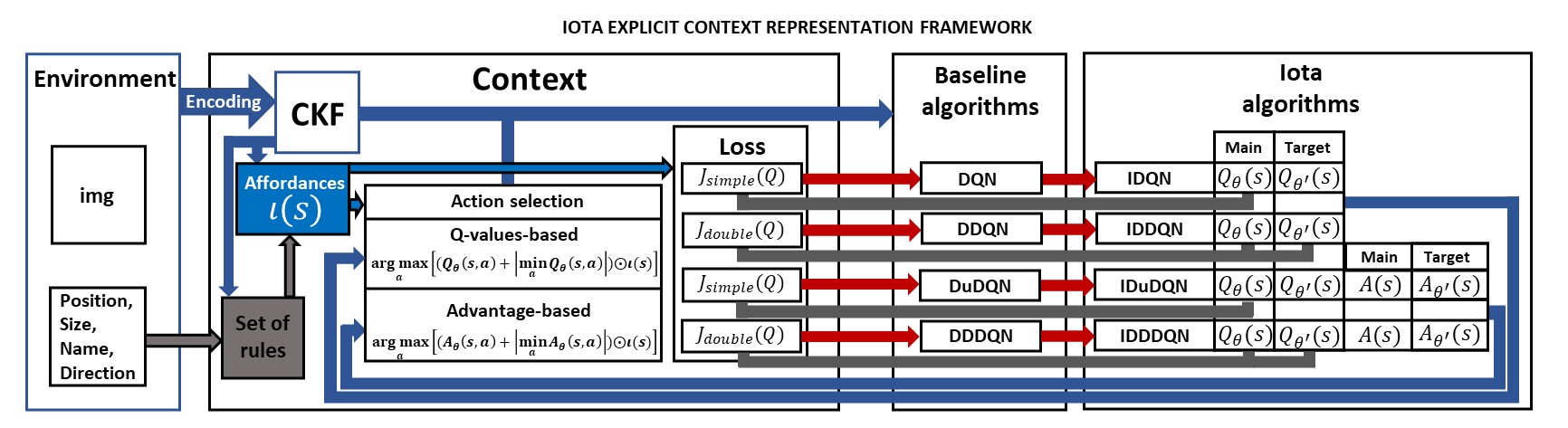}%

\caption{In this figure, the IECR framework is applied to DQN, DDQN, DuDQN, and DDQN to create four new algorithms that learn with context. The affordances function $\iota(s)$ is connected with the whole framework, hence, the name of our approach.}
\label{fig:iecr_framework}

\end{figure*}

\section{Iota Explicit Context Representation Framework}
\label{sec:iecr}

In this section, the IECR framework (Fig.~\ref{fig:iecr_framework}) is presented, aiming to allow the smooth integration of contextual information into the learning process of deep RL agents based on DQN, DDQN, DuDQN, and DDDQN. For each variant of DQN, the IECR framework uses three algorithms, detailed in the next three subsections: CKFs' generation, affordances function generation, and learning.

\begin{figure*}[!t]
\centering
\subfloat[]{\includegraphics[width=5.5in]{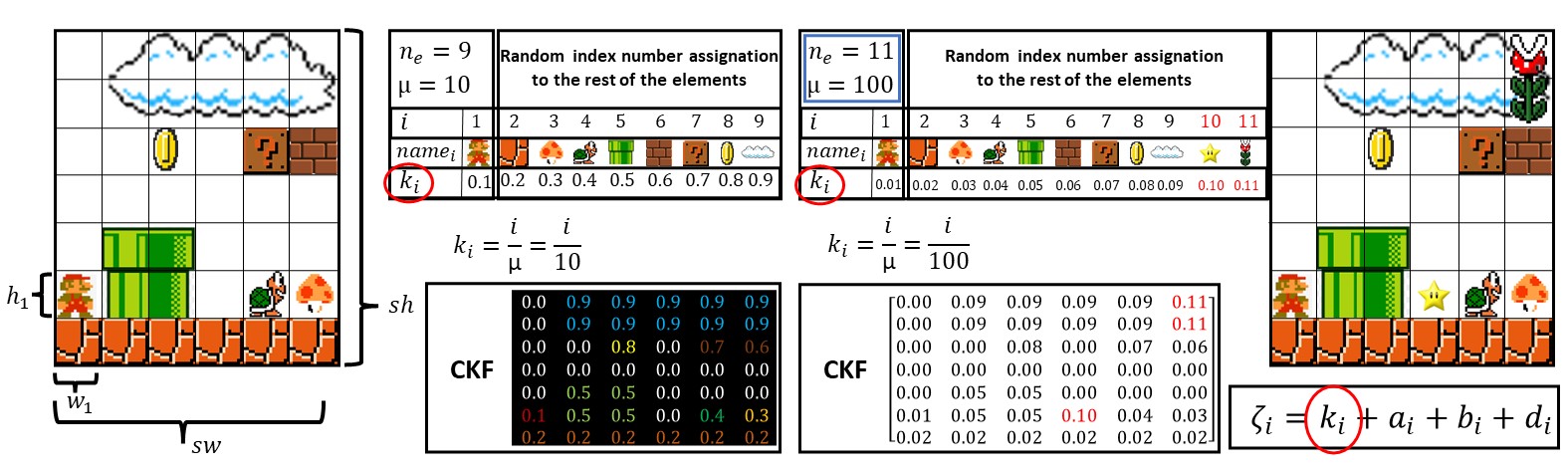}%
\label{fig:ckf_gen_a}}
\vfil
\subfloat[]{\includegraphics[width=5.5in]{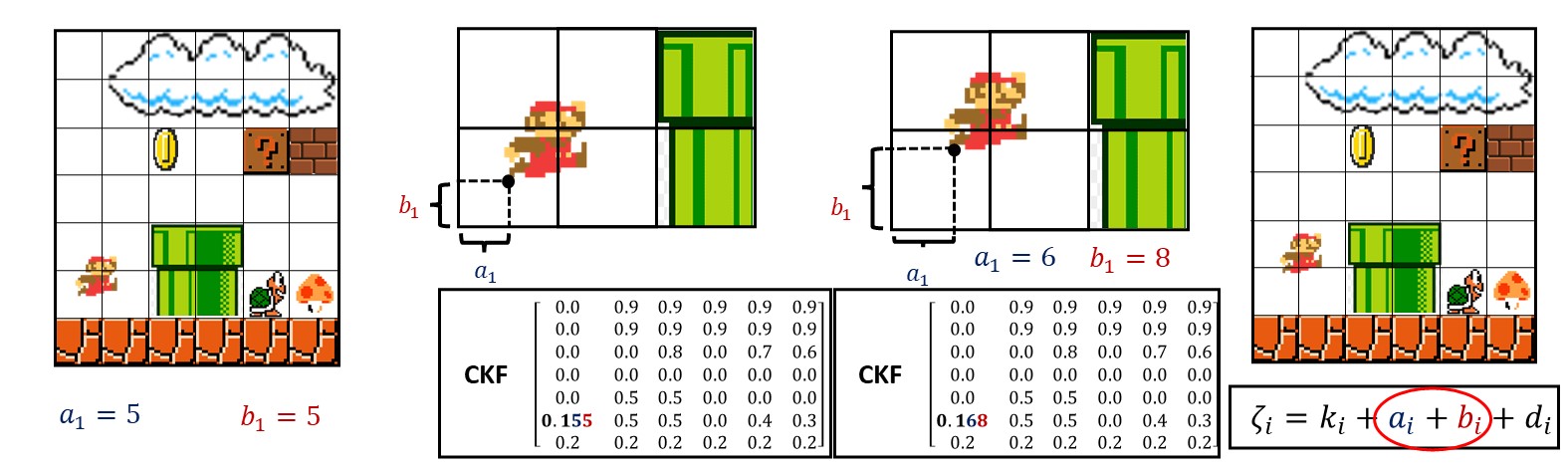}%
\label{fig:ckf_gen_b}}
\vfil
\subfloat[]{\includegraphics[width=5.5in]{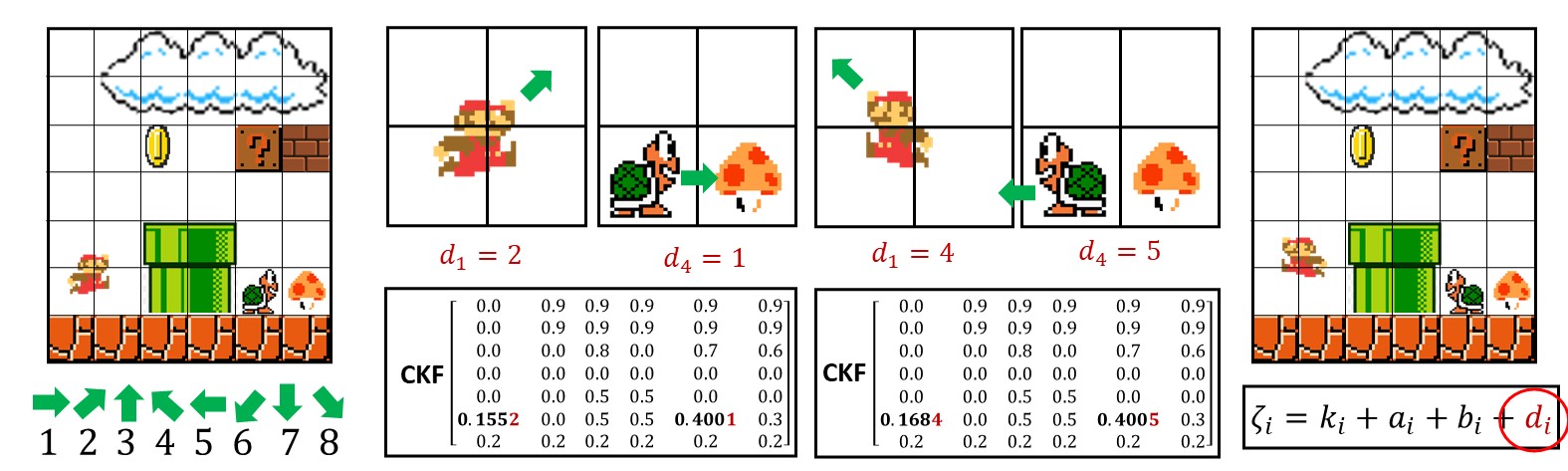}%
\label{fig:ckf_gen_c}}
\vfil

\caption{Tokenization of the state. In (a), there is an example of tokenization of the state according to the element type and the number of elements. In (b), the position of each element in the cell of the CKF is added to the token value. In (c), the direction of the elements and their numerical values are added to the token.}
\label{fig:ckf}
\end{figure*}

\subsection{Contextual key frames}

This subsection introduces Algorithm~\ref{alg:alg1}, which focuses on identifying known semantic data from the environment (objects, positions, sizes, and directions) and using it as a feature representation of the state. The contextual key frames algorithm takes as an input the semantic set $F$, the width of the screen $sw$, and its height $sh$. The output is a CKF that contains the tokens of all the elements in the environment. At this point, the goal is to create a set of tokens $Z$ that represent an $n_e$ number of elements in the environment, such that Algorithm~\ref{alg:alg1} can transform it into CKFs. The set of tokens is given by the following:

\begin{equation}\label{eq:41}
\begin{split}
Z=\{\zeta_{i}\,|\,i \in [0,n_e)\},
\end{split}
\end{equation} 

\noindent where $\zeta_i$ is the token of the $i$-th element. Formally, a CKF is a $n\times m$ matrix where CKF$_{n,m}\in Z$. In order to obtain $Z$, first, extracting the information from the environment is necessary. For this purpose, the characteristics of every component in the environment are known and tokenized such that the agent can understand them. The token $\zeta_i$ is given by: 

\begin{equation}\label{eq:40}
\begin{split}
\zeta_i=k_i+a_i+b_i+d_i,
\end{split}
\end{equation} 

\noindent where $k_i$ is the key of each type of element in the environment,  $a_i$ is the vertical numerical position, $b_i$ is the horizontal numerical position, and $d_i$ is the direction of the $i$-th element, respectively. Let:

\begin{equation}\label{eq:8}
N=\{\langle name_i,k_i\rangle\,|\,i\in E,k_i=\frac{i}{\mu}\}, 
\end{equation}

\noindent be the set of tuples representing the pair name of the element with its respective key $k_i$. Set $N$ represents the relation between the element name and its key. For example, $\langle mario,0.1\rangle$, $\langle pipe,0.3\rangle$, $\langle hole,0.8\rangle$. Let:

\begin{equation}\label{eq:9}
\begin{split}
P=\{\langle x_i,y_i,u_i,v_i,a_i,b_i \rangle\,|\,i \in [0,n_e),x_i \in [0,\infty), y_i\in[0,\infty),\\
u_i=\lfloor \frac{sw}{x_i} \rfloor, v_i=\lfloor \frac{sh}{y_i} \rfloor,
a_i=\frac{\lfloor10(\frac{sw}{x_i}-u_i)\rfloor}{10 \mu},\\
b_i=\frac{\lfloor10(\frac{sh}{y_i} -v_i )\rfloor}{100\mu}\}, 
\end{split}
\end{equation} 

\noindent be the set containing the position in the image of each element of the environment. Here, $x_i$ and $y_i$ are the horizontal and vertical positions of the agent in pixels, $u_i$ is the number of rows,  $v_i$  the number of columns, $sw$ and $sh$ are the width and height size in pixels of the screen. This set defines the values of $a_i$ and $b_i$ that depend on the token range value $\mu$. 

On the right side of Fig.~\ref{fig:ckf_gen_a}, it can be observed that the number of elements $n_e$ is equal to nine. For each element, a number of index $i$ is designated. The main element the neural network will control must have $i=1$, whereas the rest are randomly assigned. In order to calculate $k_i$, it is necessary to obtain the token range value $\mu=10^\tau$, when  $\tau\in\{1,2\}$, $10n_e \geq 10^\tau > n_e$, and $n_e<100$. This means that for the example of $n_e=9$ on the right side of the Fig.~\ref{fig:ckf_gen_a}, $\mu=10$ such that it is possible to use only one decimal to represent the key $k_i$ of the element in $\zeta_i$. When there are more than nine elements in the environment (see the left side of Fig.~\ref{fig:ckf_gen_a}), then $\mu=100$, in this manner, it is possible to represent the 11 elements with two decimals of $\zeta_i$.

Since the elements in the environment are not only static but dynamic, a method to represent their position inside the CKF is necessary. The set $P$ contains the equivalences of the width and height of every element and the position of an element inside its own grid in CKF$_{n,m}$. This idea is illustrated in Fig.~\ref{fig:ckf_gen_b}, despite the element being inside the same cell, the horizontal position $a_i$ and vertical position $b_i$ provide helpful information that can be encoded in $\zeta_i$. This representation differentiates states even when the element is situated in the same cell of the CKF.

In Fig.~\ref{fig:ckf_gen_c}, the purpose of $d_i$ is illustrated. When an element changes direction, this means a different state. Consequently, this information must be taken into account. The value of $d_i$ is in the set of movement directions $V$. Let:

\begin{equation}\label{eq:10}
\begin{split}
V=\{\langle \rightarrow,\frac{1}{1000\mu} \rangle,\langle \nearrow,\frac{2}{1000\mu} \rangle,\langle \uparrow,\frac{3}{1000\mu} \rangle,\langle \nwarrow,\frac{4}{1000\mu}
\rangle,\\ 
\langle \leftarrow,\frac{5}{1000\mu} \rangle,\langle \swarrow,\frac{6}{1000\mu} \rangle,
\langle \downarrow,\frac{7}{1000\mu} \rangle,\langle \searrow,\frac{8}{1000\mu} \rangle \},
\end{split}
\end{equation} 

\noindent be the set that contains the tuples $\langle V^1,V^2 \rangle$, where $V^1$ is the movement direction and $V^2$ its numerical value. Let:

\begin{equation}\label{eq:11}
\begin{split}
W=\{\langle w_i, h_i, \dot{w_i}, \dot{h_i}\rangle\,|\, i \in [0,n_e), w \in [0,sw],h \in [0,sh],\\
(i=1 \to \dot{w_i},\dot{h_i})\land
(i>1\to\dot{w_i}=\lceil \frac{w_i}{w_1} \rceil ,\dot{h_i}=\lceil \frac{h_i}{h_1}\rceil \},
\end{split}
\end{equation} 

\noindent be the set that represents the sizes of all the elements, where $w_i$ is the width of the element in pixels, $h_i$ is the height of the element in pixels, $\dot{w_i}$ is the width that is taken as a reference of the size of the main element of the environment, and $\dot{h_i}$ is the height of the $i$-th element taking the size of the main element of the environment as a reference. Set $W$ contains the information that represents the environment elements' sizes. With all the sets defined before, it is now possible to define the semantic set $F$, which is given by the following:

\begin{equation}\label{eq:12}
\begin{split}
F=\{\langle k_i, u_i, v_i, a_i, b_i, d_i, \dot{w_i}, \dot{h_i}\rangle\,|\, i \in [0,n_e), k_i \in N^2, \\
u_i\in P^3,v_i\in P^4,a_i\in P^5,b_i\in P^6, d_i \in V^2,\\
\dot{w_i}\in W^3, 
\dot{h_i}\in W^4\},
\end{split}
\end{equation}

\begin{algorithm}[htbp]
\caption{CKFs generator.}
\begin{algorithmic}
\STATE $\mathbf{Input:}$ Semantic set $F$, the number of elements $n_e$, the screen width $sw$, and the screen height $sh$
\STATE $\mathbf{Result:}$ CKF
\STATE $n\gets\lceil\frac{sh}{h_1}\rceil$;
\STATE $m\gets\lceil\frac{sw}{w_1}\rceil$;
\STATE Create an $n\times m$ CKF filled with $zeros$;
\STATE $\mathbf{for}$ $i=1,$ $n_e$ $\mathbf{do}$
\STATE \hspace{0.5cm}Get $k_i,u_i,v_i,a_i,b_i,d_i$ from $F$;
\STATE \hspace{0.5cm}$\mathbf{for}$ $r=0,$ $w_i$ $\mathbf{do}$
\STATE \hspace{0.5cm}\hspace{0.5cm}$\mathbf{for}$ $c=0,$ $h_i$ $\mathbf{do}$
\STATE \hspace{0.5cm}\hspace{0.5cm}\hspace{0.5cm} CKF$_{(u_i+r,v_i+c)}=\zeta_i$;
\STATE \hspace{0.5cm}\hspace{0.5cm}$\mathbf{end}$ $\mathbf{for}$
\STATE \hspace{0.5cm}$\mathbf{end}$ $\mathbf{for}$
\STATE $\mathbf{end}$ $\mathbf{for}$
\end{algorithmic}
\label{alg:alg1}
\end{algorithm}

\subsection{Iota function}

The affordances function explores the interactions among the elements of the environment. This is achieved by classifying what combinations of actions are not allowed according to the context of the environment. The affordances function is represented through  $\iota(s)$ for a given state $s$. Let:

\begin{equation}\label{eq:13}
\begin{split}
A=\{a_{j} \,|\, j \in \mathbb{Z}^{+},a\in\{0,1\} \},   
\end{split}
\end{equation} 

\noindent be the set of actions the agent can execute in the environment. Here, the total number of actions is given by $n_a=|A|$. Let:

\begin{equation}\label{eq:14}
\begin{split}
R=\{\langle a_{j},k_{i}, \phi_{i}, \alpha_{i} \rangle \,|\, a_j \in A, i \in E, \phi_i, \alpha_i \in \mathbb{Z}\},   
\end{split}
\end{equation} 

\noindent be the set of rules that contains which interactions among the actions and environment are evidently wrong, where $a$ is the action, $\phi$ is the horizontal exploration range, and $\alpha$ is the vertical exploration range. Table~\ref{table:rules} shows five sets of rules manually defined for each environment. These sets of rules contain negative affordances, which are actions that the agent is not allowed to execute given that situation.  

Algorithm~\ref{alg:alg2} takes as input the state $s$ in the shape of CKF, the set of rules $R$, the first element of the set $F$, and the number of actions $n_{a}$. All the rules related to each action are explored through the CKF obtained using Algorithm~\ref{alg:alg1}. The exploration ranges $\phi$ and $\alpha$ allow the agent to identify near or far elements that may put the agent into a bad state. However, decision-making using only $\iota(s)$  is not enough to find an optimal policy because there may be states with multiple choices where $\iota(s)$  does not provide the optimal action.

\begin{algorithm}[htbp]
\caption{$\iota(s)$ generator.}\label{alg:alg2}
\begin{algorithmic}
\STATE $\mathbf{Input:}$ CKF, number of actions $n_a$, set of rules $R$ and semantic set $F$
\STATE $\mathbf{Result:}$ $\iota(s)$
\STATE Create an $\iota$ vector with $n_{a}$ elements filled with $ones$;
\STATE Get $k_{1},u_{1},v_{1}$ from CKF;
\STATE $\mathbf{for}$ $a=0,$ $n_{a}$ $\mathbf{do}$
\STATE \hspace{0.5cm}Get $R_{a}=\langle a_{j},k_{i},\phi_{i},\alpha_{i} \rangle$ from $R$;
\STATE \hspace{0.5cm}$\mathbf{for}$ $r=0,$ $|R_a |$ $\mathbf{do}$
\STATE \hspace{0.5cm}\hspace{0.5cm}Get $k_r,\alpha_r,\alpha_r$ from $R_a$;

\STATE \hspace{0.5cm}\hspace{0.5cm}$\mathbf{for}$ $p=u_1,$ $u_1+\phi_r$ $\mathbf{do}$ 
\STATE \hspace{0.5cm}\hspace{0.5cm}\hspace{0.5cm} $\mathbf{if}$ CKF$_{p,v_1}=k_r$ $\mathbf{then}$
\STATE \hspace{0.5cm}\hspace{0.5cm}\hspace{0.5cm}\hspace{0.5cm}$\iota_a=0$;
\STATE \hspace{0.5cm}\hspace{0.5cm}$\mathbf{end}$ $\mathbf{for}$

\STATE \hspace{0.5cm}\hspace{0.5cm}$\mathbf{for}$ $l=v_1,$ $v_1+\alpha_r$ $\mathbf{do}$ 
\STATE \hspace{0.5cm}\hspace{0.5cm}\hspace{0.5cm} $\mathbf{if}$ CKF$_{u_1,l}=k_r$ $\mathbf{then}$
\STATE \hspace{0.5cm}\hspace{0.5cm}\hspace{0.5cm}\hspace{0.5cm}$\iota_a=0$;
\STATE \hspace{0.5cm}\hspace{0.5cm}$\mathbf{end}$ $\mathbf{for}$
\STATE \hspace{0.5cm}$\mathbf{end}$ $\mathbf{for}$
\STATE $\mathbf{end}$ $\mathbf{for}$
\STATE $\iota(s)\gets \iota$;
\end{algorithmic}
\end{algorithm}

\subsection{Learning}

This subsection explains how the CKFs and $\iota(s)$ are incorporated into the DQN, DDQN, DuDQN, and DDDQN algorithms. For this purpose, we introduce Algorithm~\ref{alg:alg3} and how it can be applied to DQN variants to create the four new algorithms developed in this research: IDQN, IDDQN, IDuDQN, and IDDDQN. The main differences (Fig.~\ref{fig:iecr_framework}) among the algorithms are in their neural network architecture (single or dueling), action selection (Q-values-stream-based or Advantage-stream-based), and loss functions (simple or double).

\begin{algorithm}
\caption{Context-based learning}\label{alg:alg3}
\begin{algorithmic}
\STATE Set neural network architecture;
\STATE \textbf{$\backslash\ast$} Single stream for IDQN and IDDQN, and double stream for IDuDQN and IDDDQN \textbf{$\ast\backslash$}
\STATE Initialize main and target neural networks with random weights $\theta$ and $\theta'$, respectively;\\

\STATE Initialize a buffer $D^{replay}$;

\STATE$\mathbf{for}$ $n$ number of episodes $\mathbf{do}$

\STATE\hspace{0.5cm}$\mathbf{for}$ $t=1,$ $T$ $\mathbf{do}$

\STATE\hspace{0.5cm}\hspace{0.5cm}Get a CKF with Algorithm~\ref{alg:alg1};

\STATE\hspace{0.5cm}\hspace{0.5cm}Get $\iota(s)$ with Algorithm~\ref{alg:alg2};

\STATE \hspace{0.5cm}\hspace{0.5cm}With probability $\epsilon$, execute a valid action;\\
\STATE\hspace{0.5cm}\hspace{0.5cm}\textbf{$\backslash\ast$} Eq.~\eqref{eq:15} for IDQN and IDDQN or Eq.~\eqref{eq:24} 
\STATE\hspace{0.5cm}\hspace{0.5cm}IDuDQN and IDDDQN \textbf{$\ast\backslash$}

\STATE \hspace{0.5cm}\hspace{0.5cm}Store $s,s',r,\iota(s),\iota(s')$ transition  in $D^{replay}$;

\STATE \hspace{0.5cm}\hspace{0.5cm}Sample a transition from $D^{replay}$;

\STATE \hspace{0.5cm}\hspace{0.5cm}$\mathbf{if}$ the state is terminal $\mathbf{then}$

\STATE \hspace{0.5cm}\hspace{0.5cm}\hspace{0.5cm}Set $J_{\iota}(Q)=0$ \\

\STATE \hspace{0.5cm}\hspace{0.5cm}\hspace{0.5cm}\textbf{$\backslash\ast$} $target=r$ for IDQN and IDuDQN or
\STATE \hspace{0.5cm}\hspace{0.5cm}\hspace{0.5cm}$target_{double}=r$ for IDDQN and IDDDQN \textbf{$\ast\backslash$}

\STATE \hspace{0.5cm}\hspace{0.5cm}$\mathbf{else}$ 

\STATE \hspace{0.5cm}\hspace{0.5cm}\hspace{0.5cm} Calculate the loss;\\
\STATE \hspace{0.5cm}\hspace{0.5cm}\hspace{0.5cm}\textbf{$\backslash\ast$} $J_{simple}(Q)$ for IDQN and IDuDQN or
\STATE \hspace{0.5cm}\hspace{0.5cm}\hspace{0.5cm}$J_{double}(Q)$ for IDDQN and IDDDQN \textbf{$\ast\backslash$}

\STATE \hspace{0.5cm}\hspace{0.5cm}Perform gradient descent step on  the loss;

\STATE \hspace{0.5cm}\hspace{0.5cm}Update $\theta$; 

\STATE \hspace{0.5cm}\hspace{0.5cm}$\mathbf{if}$ $t\mod\tau=0$ $\mathbf{then}$

\STATE \hspace{0.5cm}\hspace{0.5cm}\hspace{0.5cm}$\theta'\gets\theta$

\STATE \hspace{0.5cm}\hspace{0.5cm}$s\gets s'$

\STATE \hspace{0.5cm}$\mathbf{end}$ $\mathbf{for}$

\STATE $\mathbf{end}$ $\mathbf{for}$

\end{algorithmic}
\end{algorithm} 

\begin{figure*}[!t]
\centering
\subfloat[]{\includegraphics[width=3.2in]{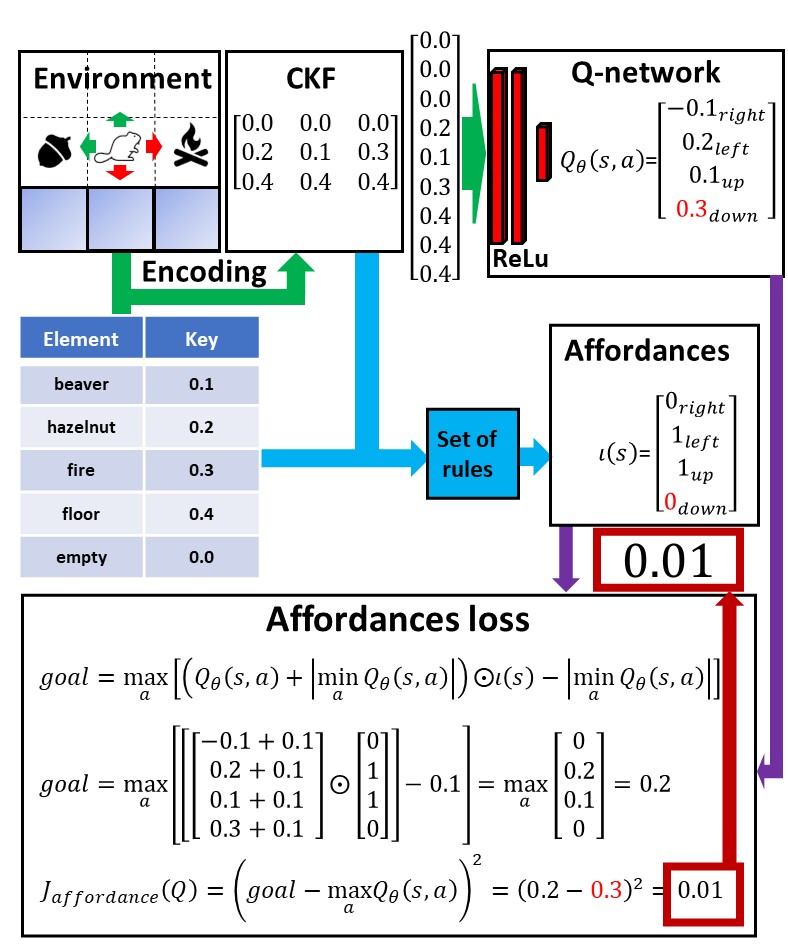}%
\label{fig:affordance_loss_a}}
\hfil
\subfloat[]{\includegraphics[width=3.2in]{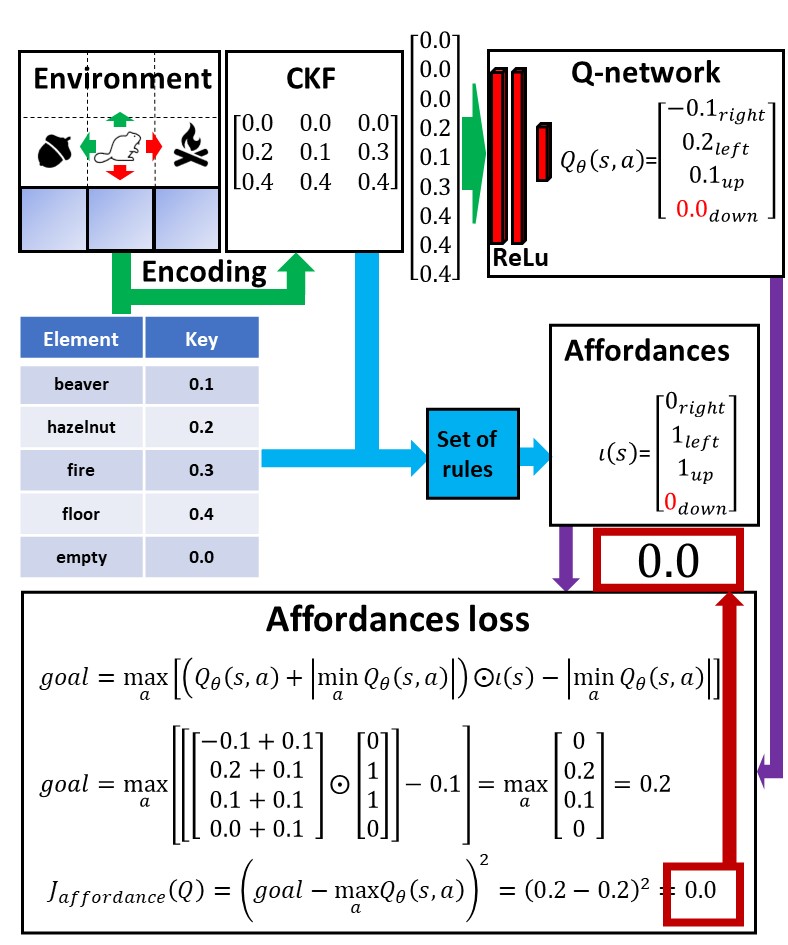}%
\label{fig:affordance_loss_b}}
\caption{This figure illustrates how the affordances can be used to influence neural networks' learning. In (a), the neural network predicts a non-valid action. Consequently, the affordances loss increases. In (b), the neural network outputs respect the boundaries given by $\iota(s)$. This provokes the affordances loss to be equal to zero.}
\label{fig:affordance_loss}
\end{figure*}

We begin with \textbf{IDQN}, which is the implementation of the IECR framework into DQN. Then, the affordances function $\iota(s)$, used for selection-making tasks, is given by the following:

\begin{equation}\label{eq:15}
\begin{split}
a_t=\argmax_{a}[(Q_{\theta}(s,a)+|\min_{a}Q_{\theta}(s,a)|)\odot\iota(s)]
\end{split}
\end{equation} 

Adding $\iota(s)$ to the action selection process produces useful data but not optimal solutions. Hence, it is necessary to include $\iota(s)$ into the DQN learning Eq.~\eqref{eq:2} by applying the Hadamard product operation:

\begin{equation}\label{eq:16}
\begin{split}
target=r+\gamma\max_{a'} [(Q_{\theta'}(s',a')+\\
|\min_{a'}Q_{\theta'}(s',a')|)\odot\iota(s')-\min_{a'}Q_{\theta'}(s',a')]
\end{split}
\end{equation} 

For IDQN, the main simple neural network loss $J_{\iota n}(Q)$  is given by the following:

\begin{equation}\label{eq:17}
\begin{split}
J_{\iota n}(Q)=(target-Q_{\theta}(s,a))^2                       \end{split}
\end{equation}

Since it is possible to obtain $\iota(s)$, the target value goal for the main neural network, which indicates how the network overestimates impossible actions according to $\iota(s)$, can be calculated as follows:

\begin{equation}\label{eq:18}
\begin{split}
goal=r+\gamma\max_{a} [(Q_{\theta}(s,a)+\\
|\min_{a}Q_{\theta'}(s,a)|)\odot\iota(s)-\min_{a}Q_{\theta}(s,a)]
\end{split}
\end{equation}                                     

In Fig.~\ref{fig:affordance_loss}, the affordance loss $J_{affordance}(Q)$ functionality is illustrated. The loss increases when the neural network overestimates a non-valid action (Fig.~\ref{fig:affordance_loss_a}). On the contrary, when the neural network respects the constraints given by $i(s)$, the loss is equal to zero (Fig.~\ref{fig:affordance_loss_b}). The affordance loss is given by the following:

\begin{equation}\label{eq:19}
\begin{split}
J_{affordance}(Q)=(goal-\max_{a}Q_{\theta}(s,a))^2
\end{split}
\end{equation} 

The total simple loss $J_{simple}(Q)$  is the sum of the main simple neural network loss $J_{\iota n}(Q)$ and the affordance loss $J_{affordance}(Q)$:

\begin{equation}\label{eq:20}
\begin{split}
J_{simple}(Q)=J_{\iota n}(Q)+\lambda J_{affordance}(Q)
\end{split}
\end{equation}

The $\lambda$ parameter controls the effect between losses. We examine removing the affordance loss in the next section. Algorithm~\ref{alg:alg3}. \textbf{IDDQN} is the implementation of DDQN in IDQN. The difference when compared with IDQN is in its $target_{double}$ equation given by the following:

\begin{equation}\label{eq:21}
\begin{split}
target_{double}=r+\gamma Q_{\theta'}(s',\argmax_{a'}[(Q_{\theta'}(s',a')+\\
|\min_{a'}Q_{\theta'}(s',a')|)\odot\iota(s')-\min_{a'}Q_{\theta}(s',a')])
\end{split}
\end{equation}

Given $target_{double}$ is possible to obtain the main double neural network loss $J_{\iota nd}(Q)$:

\begin{equation}\label{eq:22}
\begin{split}
J_{\iota nd}(Q)=(target_{double} - \max_{a}Q{\theta}(s,a))^2
\end{split}
\end{equation} 

The total double loss for IDDQN $J_{double}(Q)$  is the sum of the main double neural network loss $J_{\iota nd}(Q)$  and the affordance loss $J_{affordance}(Q)$:

\begin{equation}\label{eq:23}
\begin{split}
J_{double}(Q)=J_{\iota nd}(Q)+\lambda J_{affordance}(Q)
\end{split}
\end{equation}

IDDQN differs from IDQL in the way the total loss is calculated. IDQL uses $J(Q)_{simple}$, whereas IDDQN uses the double loss $J(Q)_{double}$. For action selection, both algorithms use Eq.~\eqref{eq:15}. IDQN and IDDQN use neural networks with single streams (Fig.~\ref{fig:sta_architectures} top image)

On the other hand, dueling architectures use the advantage stream to select an action during the training process. \textbf{IDuDQN} utilizes neural networks with dueling streams (Fig.~\ref{fig:sta_architectures} bottom image) and the simple loss function $J(Q)_{simple}$. However, the DuDQN action selection equation is given by the following:

\begin{equation}\label{eq:24}
\begin{split}
a_t=\argmax_{a}[(A_{\theta}(s,a)+|\min_{a}A_{\theta}(s,a)|)\odot\iota(s)]
\end{split}
\end{equation} 

\textbf{IDDDQN} utilizes neural networks with dueling streams (Fig.~\ref{fig:sta_architectures} bottom image), the first for the advantage and the second for the Q-values. It also selects an action from the advantage stream using the Eq.~\eqref{eq:24}. DDDQN uses the double loss $J(Q)_{double}$. 

IDQN, IDDQN, IDuDQN, and IDDDQN require a buffer replay $D^{replay}$, a main neural network $Q_\theta$, and a target neural network $Q_\theta'$. The weights of the main neural network $\theta$ are copied to the target neural network weight $\theta'$ every $n_{step}$ steps, where $n_{step}$ is usually set to 100. The pseudo-code of Algorithm~\ref{alg:alg3} explains how to implement the algorithms mentioned above.

\section{Experimental setup}
\label{sec:setup}

\begin{figure}[htbp]
    \centering
    \smallskip
    \begin{tabular}{ c c c}
        \begin{tabular}{c}
        \smallskip
            \subfloat[]{\includegraphics[width=0.9in]{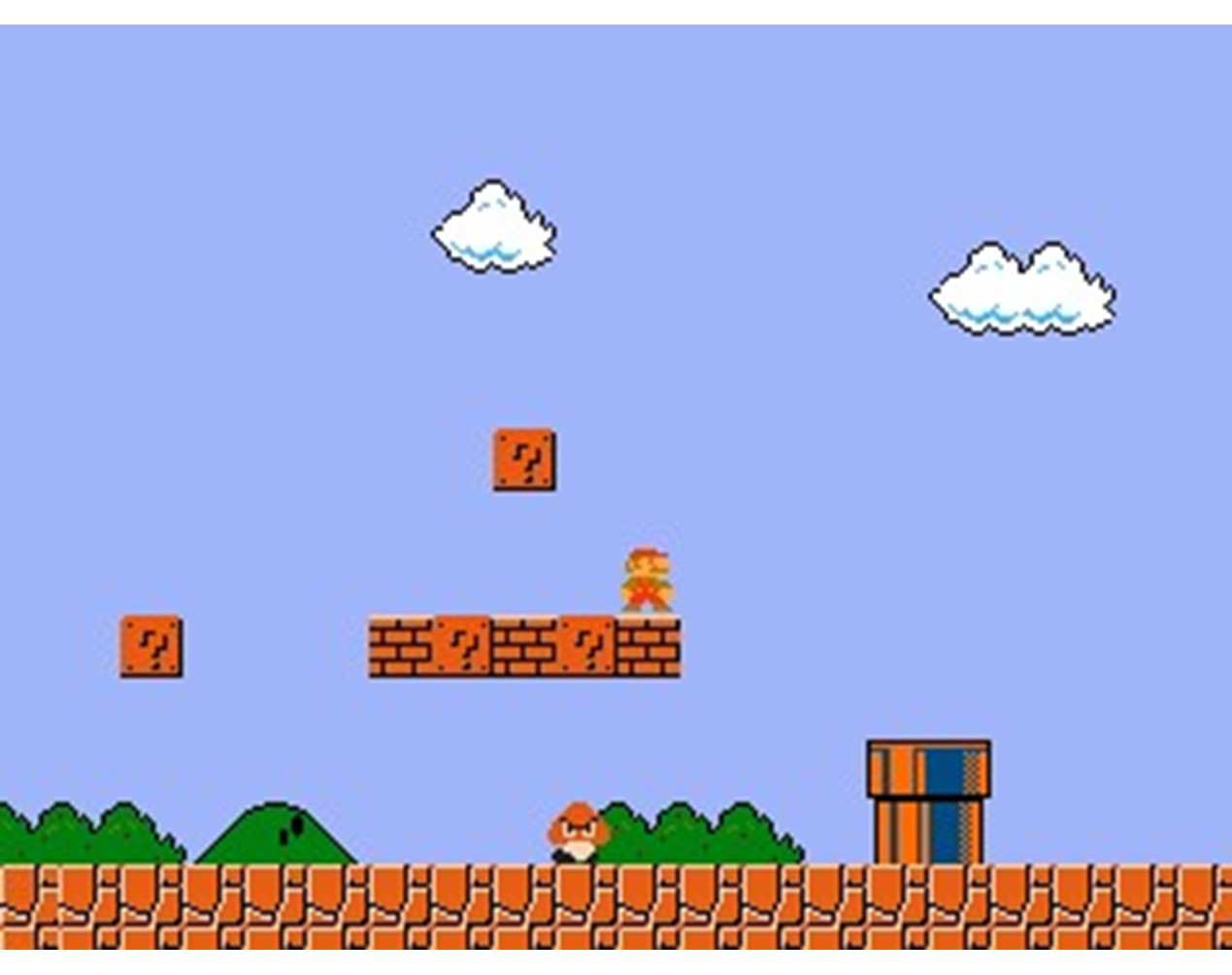}%
            \label{fig:env_mario}}\\
            \subfloat[]{\includegraphics[width=0.9in]{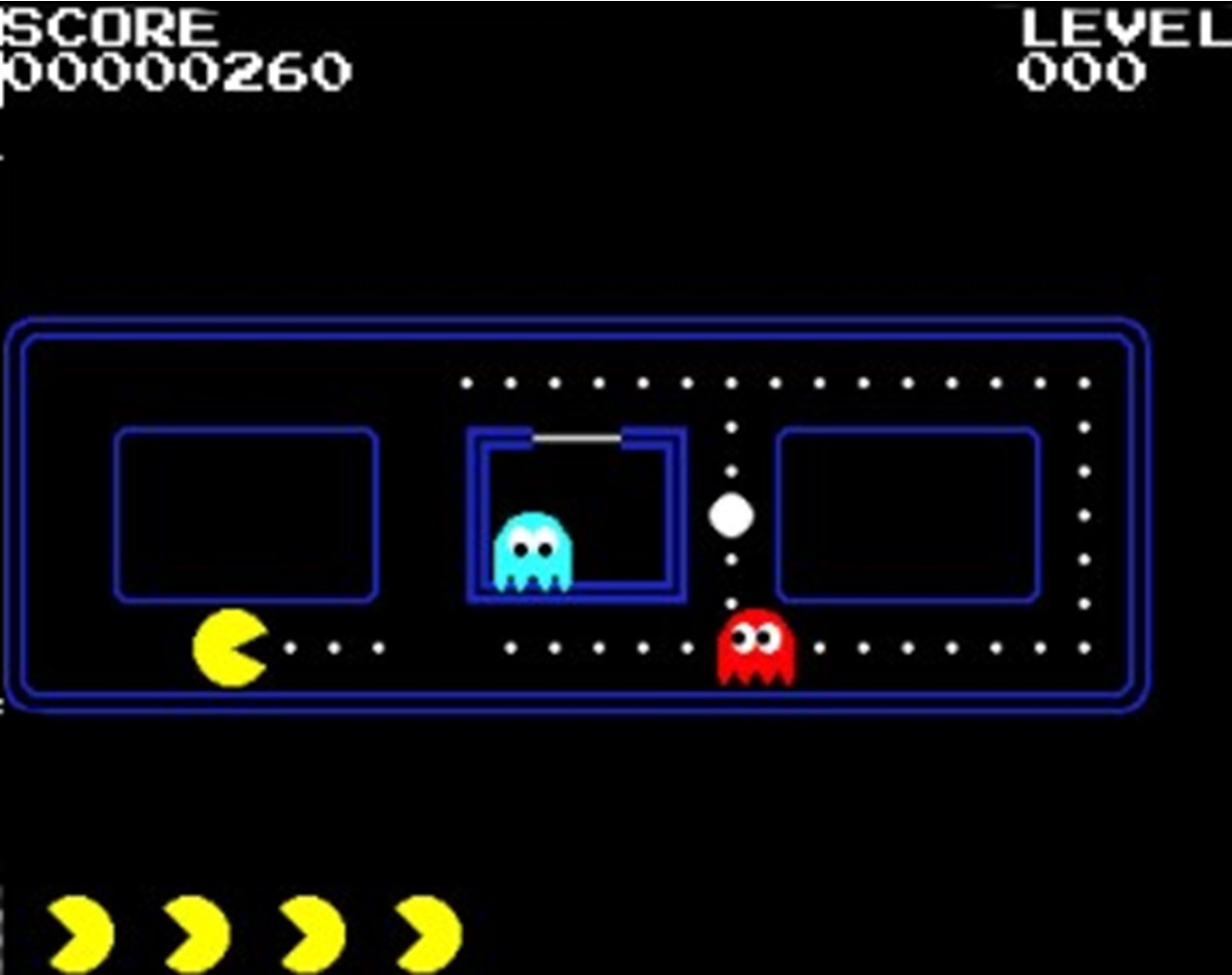}%
            \label{fig:env_pacman}}
        \end{tabular}
        &
        \begin{tabular}{c}
        \smallskip
            \subfloat[]{\includegraphics[width=0.8in]{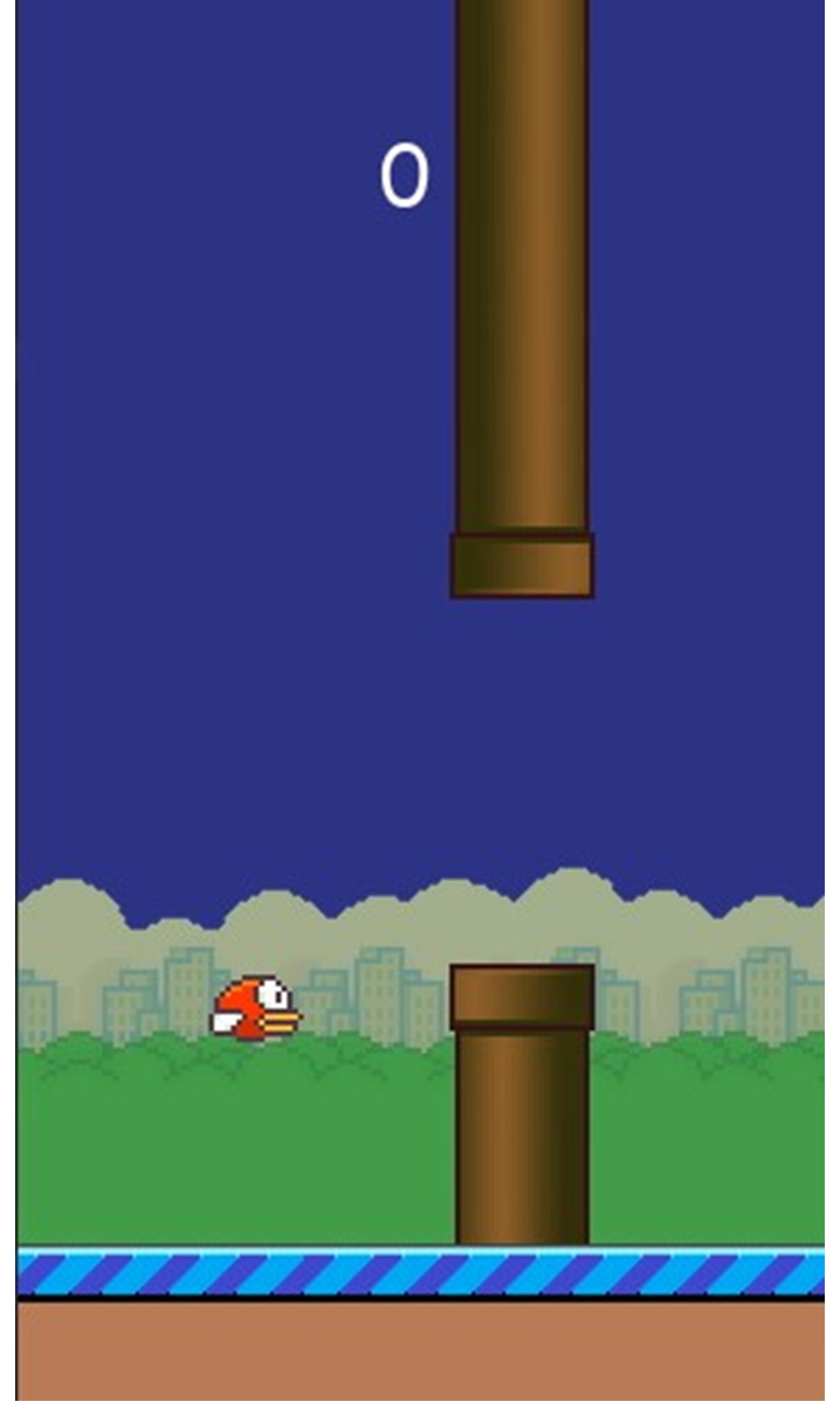}%
            \label{fig:env_flappybirds}}
        \end{tabular}
        &
        \begin{tabular}{c}
        \smallskip
            \subfloat[]{\includegraphics[width=0.6in]{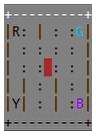}%
            \label{fig:env_taxi}}\\
            \subfloat[]{\includegraphics[width=1.0in]{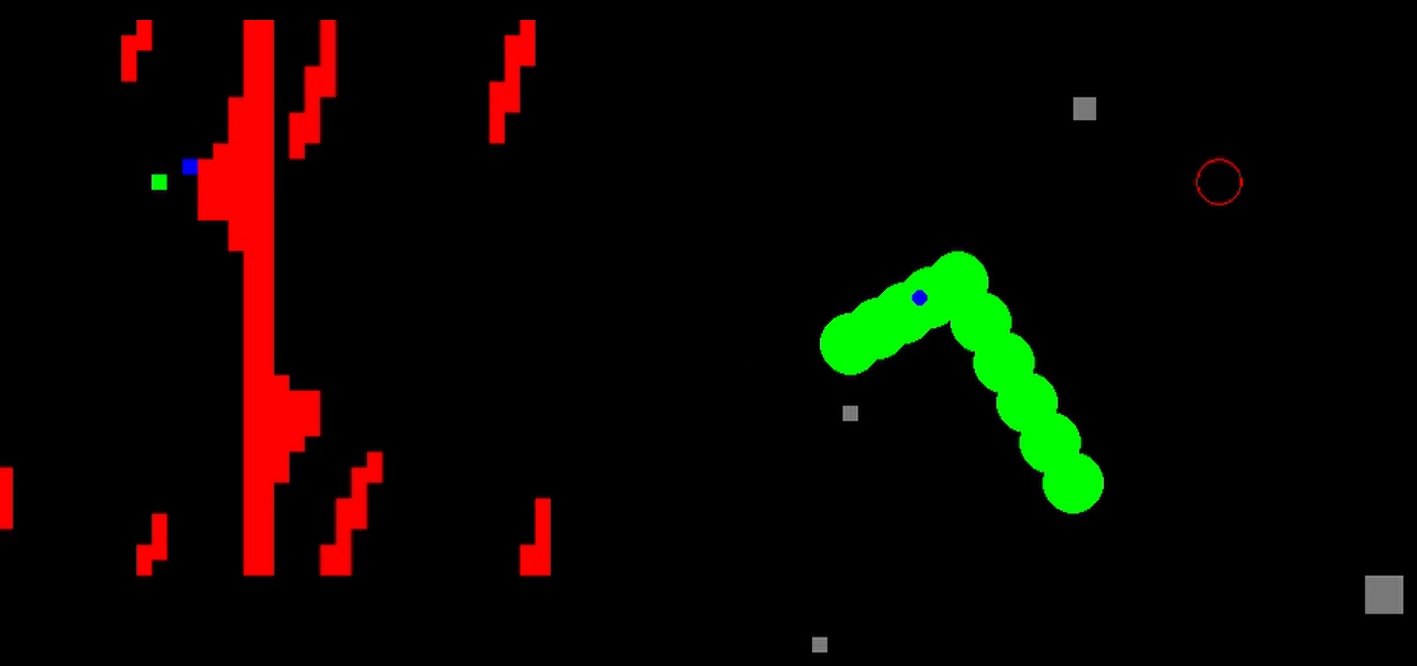}%
            \label{fig:env_scara}}
        \end{tabular}
    \end{tabular}
    \caption{Game environments. (a) Mario. (b) Pacman. (c) FlappyBirds. (d) TaxiDriver. (e) ScaraRobot.}
    \label{fig:env}
\end{figure}

This section describes the experimental setup used in this paper to evaluate the performance of the IDQN, IDDQN, IDuDQN, and IDDDQN algorithms. For our experiments, we utilized five environments (Fig.~\ref{fig:env}). Each environment has distinctive characteristics that challenge the algorithms in multiple ways. The experiments are conducted in two stages. The first stage of experiments evaluates how IDQN performs against DQN and a random policy to prove that IDQN can learn from the CKFs' representation. The second stage of experiments compares the performance of IDQN, IDDQN, IDuDQN, and IDDDQN against their state-of-the-art variants. For both stages of experiments, the CKFs' representation is used as the input of the neural networks. To compare the performance of our algorithms, we used DQN, DDQN, DuDQN, DDDQN, PPO, and A2C implementations of the stable-baselines. 

\begin{figure}[!t]
\centering
\subfloat[]{\includegraphics[width=3.5in]{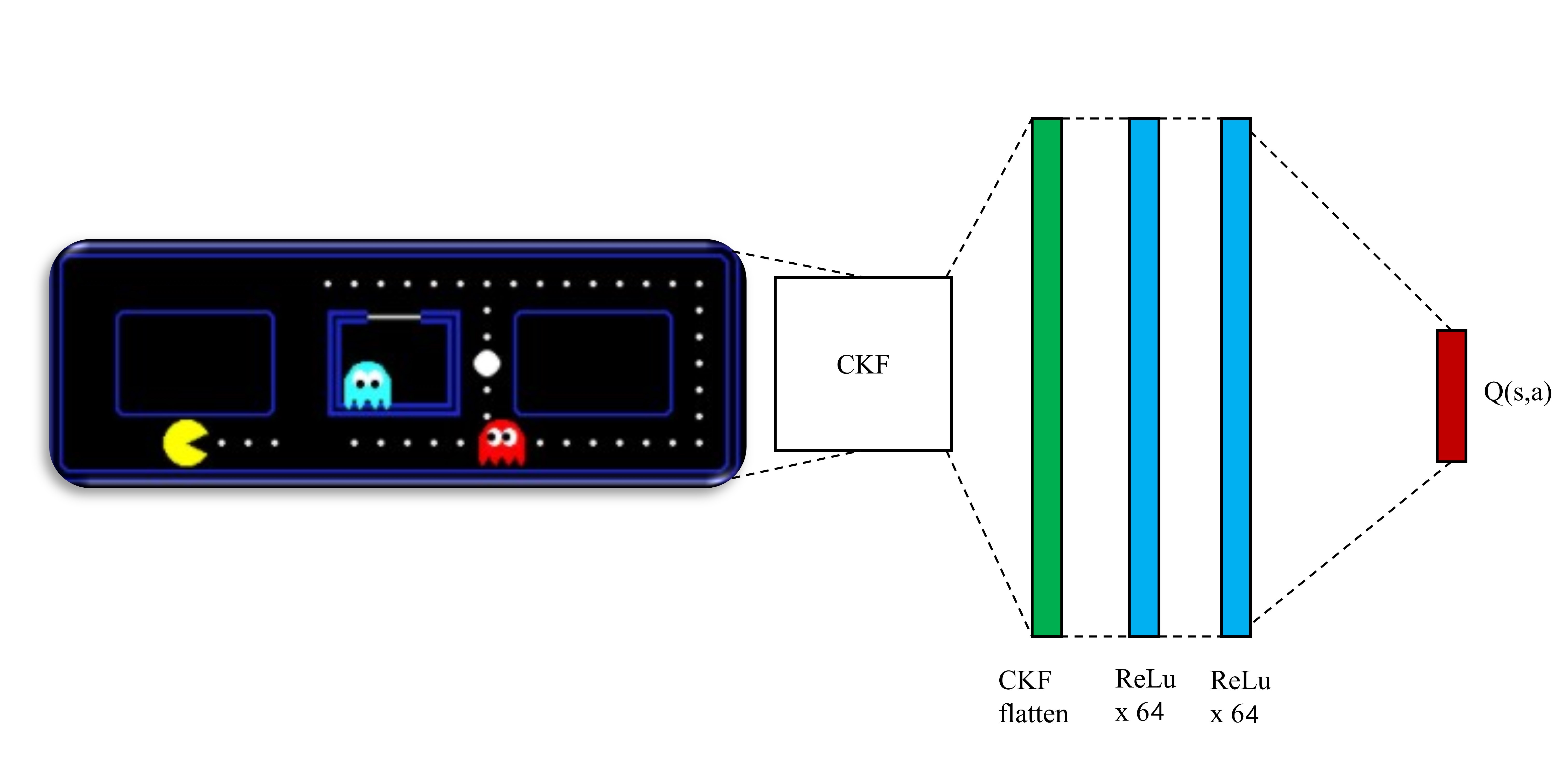}%
\label{fig:architecture_1}}
\hfil
\subfloat[]{\includegraphics[width=3.5in]{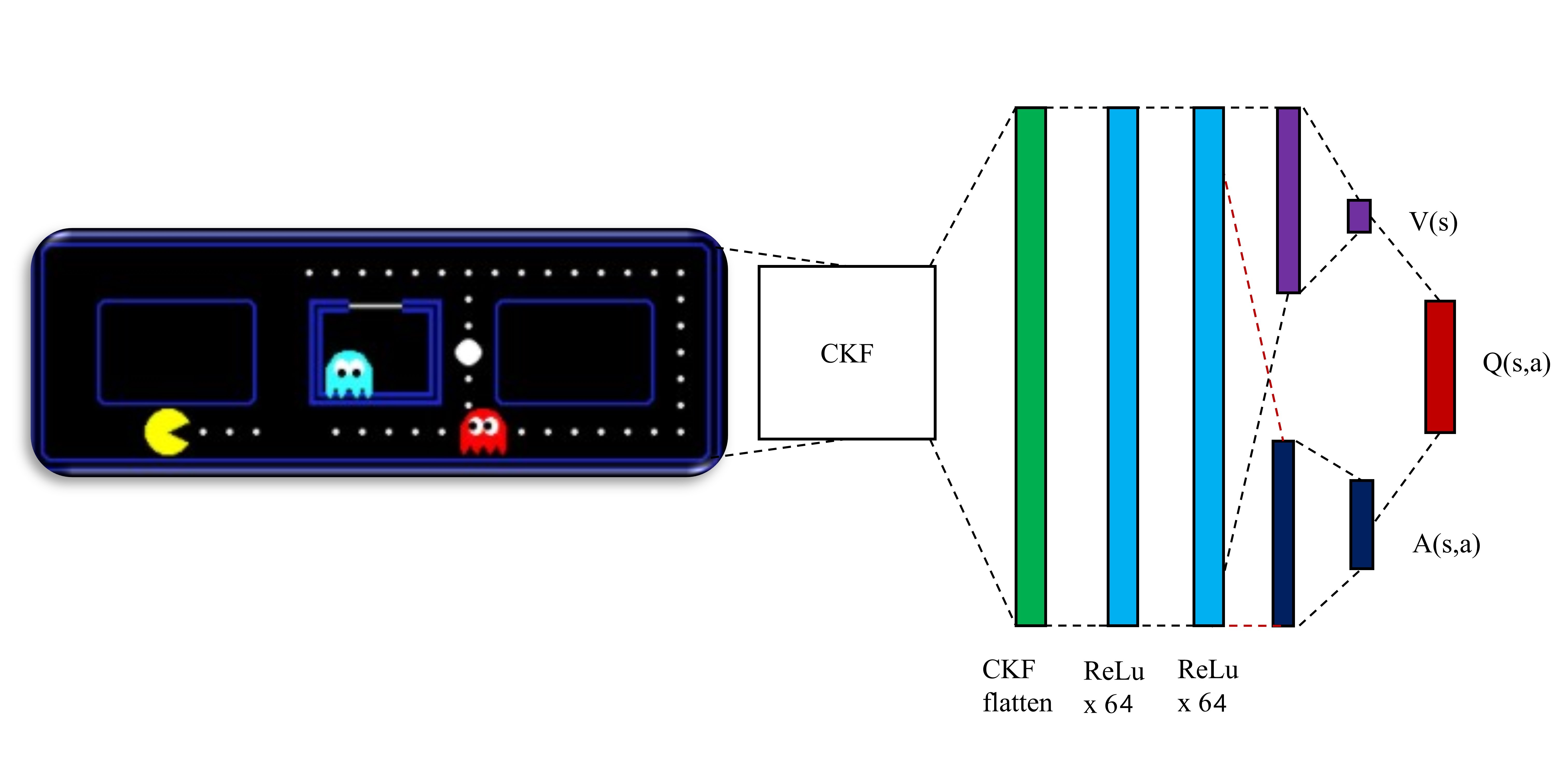}%
\label{fig:architecture_2}}
\caption{Neural network architectures used in the experiments. (a) Neural network setup used for IDQN and IDDQN. (b) Neural network setup used by  IDuDQN and IDDDQN.}
\label{fig:architecture}
\end{figure}

In the first stage, we measured the learning progress in every episode by using the average reward. This calculation is based on the decisions of the neural network. In the second stage, the progress was measured in epochs. We used the same network architecture and hyperparameters for all the experiments. The neural network architectures (single and double stream) are shown in Fig.~\ref{fig:architecture}. To this end, Table~\ref{table:parameters} shows the parameter values used for all the environments and experiments.

\begin{table}
\begin{center}
\caption{\label{table:parameters}Parameters used during the experiments.}
\begin{tabular}{ |p{0.3\textwidth}|p{0.1\textwidth}| }
 \hline
 Parameter & Value \\
 \hline
Learning rate ($\gamma$) & $0.99$\\
 \hline
 Mini-batch size & $64$\\ 
 \hline
 Episodes & $2000$\\ 
 \hline
 Epochs  & $200$\\ 
  \hline
 Optimizer  & Adam\\ 
 \hline
 Optimizer learning rate & $0.0001$\\ 
 \hline
 Loss function & $huber$ \\ 
 \hline
 Training steps per epoch & $400$\\ 
 \hline
 Maximum steps per episode & $3000$\\ 
 \hline
 Target network update steps($\tau$) & $100$\\ 
 \hline
 Replay buffer size & $50000$ \\ 
 \hline
 $\epsilon_{max}$ & 0.9 \\ 
 \hline
 $\epsilon_{min}$ & 0.05 \\ 
 \hline
 $\sigma_{episode}$ ($\epsilon$ decay per episode) & 0.001 \\ 
 \hline
 $\sigma_{epoch}$ ($\epsilon$ decay per epoch) & 0.01 \\ 
 \hline
\end{tabular}
\end{center}
\end{table}

\subsection{Environments description}

Mario (Fig.~\ref{fig:env_mario}) is a game in which the free space available in the environment allows the agent to move almost everywhere, and this is a challenge. Moreover, negative rewards could take longer to propagate. For example, suppose there is a hole, and the jumping action was executed several states before. In that case, it will take several steps to make the agent understand that the action taken puts the agent in a deficient state. The reward function of the Mario environment is given by:

\begin{equation}\label{eq:25}
\begin{split}
    r_{mario}(s,a)= 
\begin{cases}
    10,& \text{finishing the game}\\
    -10,&\text{dying}\\
    1,&\text{moving forward one square}\\
    0,&\text{otherwise}
\end{cases}
\end{split}
\end{equation}

Pacman (Fig.~\ref{fig:env_pacman}) is a simple game where the agent's path is highly constrained. An action that immediately crashes with a wall or an enemy is easily identifiable in this environment. However, when using the $\epsilon$-greedy policy, the agent does not consider evident mistakes. On the other hand, the $\iota(s)$ function can avoid immediate mistakes and accelerate the agent's learning. The reward function of the Pacman environment is given by:

\begin{equation}\label{eq:26}
\begin{split}
    r_{pacman}(s,a)= 
\begin{cases}
    10,& \text{finishing the game}\\
    -10,&\text{dying}\\
    1,&\text{eating a pellet}\\
    0,&\text{otherwise}
\end{cases}
\end{split}
\end{equation}

Even though FlappyBirds (Fig.~\ref{fig:env_flappybirds}) is a game with only two actions (fly and fall), their random selection under the same probability provokes the agent to stay at the top of the screen, since the fly action has a more biased behavior than the fall action. The replay buffer will eventually fill with useless data that affect the agent's learning process. The reward function of the FlappyBirds environment is given by:

\begin{equation}\label{eq:27}
\begin{split}
    r_{flappybirds}(s,a)= 
\begin{cases}
    10,& \text{finishing the game}\\
    -10,&\text{dying}\\
    1,&\text{passing the pipes}\\
    0,&\text{otherwise}
\end{cases}
\end{split}
\end{equation} 

TaxiDriver (Fig.~\ref{fig:env_taxi}) and ScaraRobot (Fig.~\ref{fig:env_scara}) are environments that can only be solved by finishing two tasks: pick and drop. This characteristic complicates the exploration-exploitation process of the agent because the buffer fills more with demonstrations of the first task (pick) than the second one (drop). The reward function of the TaxiDriver environment is given by:

\begin{equation}\label{eq:28}
\begin{split}
    r_{taxidriver}(s,a)= 
\begin{cases}
    10,& \text{picking in the right place}\\
    10,& \text{dropping in the right place}\\
    0,&\text{otherwise}
\end{cases}
\end{split}
\end{equation} 

The reward function of the ScaraRobot environment is given by the following:

\begin{equation}\label{eq:29}
\begin{split}
    r_{ScaraRobot}(s,a)= 
\begin{cases}
    10,& \text{picking in the right place}\\
    10,& \text{dropping in the right place}\\
    1,&\text{getting closer to the goal}\\
    0,&\text{otherwise}
\end{cases}
\end{split}
\end{equation} 

For all our experiments, we took the average reward (produced from the actions of the neural network) as an indicator of learning progress. The set of rules for each environment is given in Table~\ref{table:rules}.

\begin{table}[htbp]
\centering
\caption{Set of rules for each game environment.\label{table:rules}}
\begin{tabular}{ |p{0.1\textwidth}|p{0.3\textwidth}| }
\hline
Environment & Set of rules \\
\hline
Mario & $R_{Mario}=\{\langle right,pipe,1,0 \rangle,$
 
$\langle right, enemy,2,0 \rangle,$

$\langle right, hole,2,0 \rangle, $

$\langle right, block,1,0 \rangle, $

$\langle jump,empty,0,-1 \rangle\}$ \\ 
 \hline
Pacman & $R_{Pacman}=\{\langle right,wall,1,0 \rangle, $
$\langle left, wall,-1,0 \rangle, $

$\langle up, wall,0,1 \rangle, $

$\langle down,wall,0,-1 \rangle,$

$ \langle right, ghost,1,0 \rangle, $

$\langle left, ghost,-1,0 \rangle,$

$ \langle up, ghost,0,1 \rangle, $

$\langle down,ghost,0,-1 \rangle \}$

\\ 
 \hline
 FlappyBirds & $R_{FlappyBirds}=\{\langle fly, pipe_{up} , 3,0 \rangle,$
 
 $\langle fly, pipe_{up} , 0 , 1 \rangle, $
 
 $\langle fly, ceiling, 0,1 \rangle,$
 
 $\langle fall, pipe_{down} , 3,0 \rangle, $
 
 $\langle fall, pipe_{down} , 0 , -1 \rangle,$ 
 
 $\langle fall, floor , 0 , -1 \rangle\}$  \\ 
 \hline
 TaxiDriver & $R_{TaxiDriver}=\{\langle right,wall,1,0 \rangle, $
$\langle left, wall,-1,0 \rangle, $
$\langle up, wall,0,1 \rangle, $
$\langle down,wall,0,-1 \rangle,$
$ \langle pick, wall,0,0 \rangle, $
$ \langle pick, empty,0,0 \rangle, $
$ \langle drop, wall,0,0 \rangle, $
$\langle drop,empty,0,0 \rangle \}$  \\ 
 \hline
ScaraRobot & $R_{ScaraRobot}=\{\langle right,obstacle,1,0 \rangle, $
$\langle left, obstacle,-1,0 \rangle, $
$\langle up, obstacle,0,1 \rangle, $
$\langle down,obstacle,0,-1 \rangle,$
$ \langle pick, obstacle,0,0 \rangle, $
$ \langle pick, empty,0,0 \rangle, $
$ \langle drop, obstacle,0,0 \rangle, $
$\langle drop,empty,0,0 \rangle \}$  \\  
 \hline
\end{tabular}
\end{table}

\subsection{First stage of experiments}

This experiment aimed to prove the neural networks' capacity to learn from the CKFs. During this first stage of the experiments, we performed IDQN for the five environments and utilized the number of episodes as a common reference. Every episode is constrained to end if the agent reaches 3,000 steps, dies, or completes the game. In addition, we ran three extra experiments for each environment. Thus, this could confirm that IDQN performs better than DQN and a full random policy. We also performed DQN using the same number of episodes IDQN took to learn.

\subsection{Second stage of experiments}

A problem with the first stage is that the incorporation of the affordance function $\iota(s)$ allows the agent to last longer during every episode. However, DQN makes more mistakes, and the number of training steps for the neural networks is significantly lower because every episode ends prematurely. The goal of this second stage is to deal with this unfair comparison. For this purpose, we measure the learning progress in epochs, where one epoch is conformed of 400 training steps. To this end, we also compared the influence of the affordance loss Eq. \eqref{eq:19} by setting $\lambda$ to 0, 0.5, 1, 5 and 10, respectively.

\section{Results}
\label{sec:results}
During this section, we describe the learning curves in Fig.~\ref{fig:results} from the results of the two stages of the experiments. The affordance loss impact is also discussed at the end of this section. Demo available at \url{https://youtu.be/Gqsud7KUZfM}.

\subsection{Results of the first stage of the experiments}

\begin{figure}[htbp]
\centering
\smallskip
\begin{tabular}{ c c }
        \begin{tabular}{c}
        \smallskip
        \subfloat[]{\includegraphics[width=1.5in]{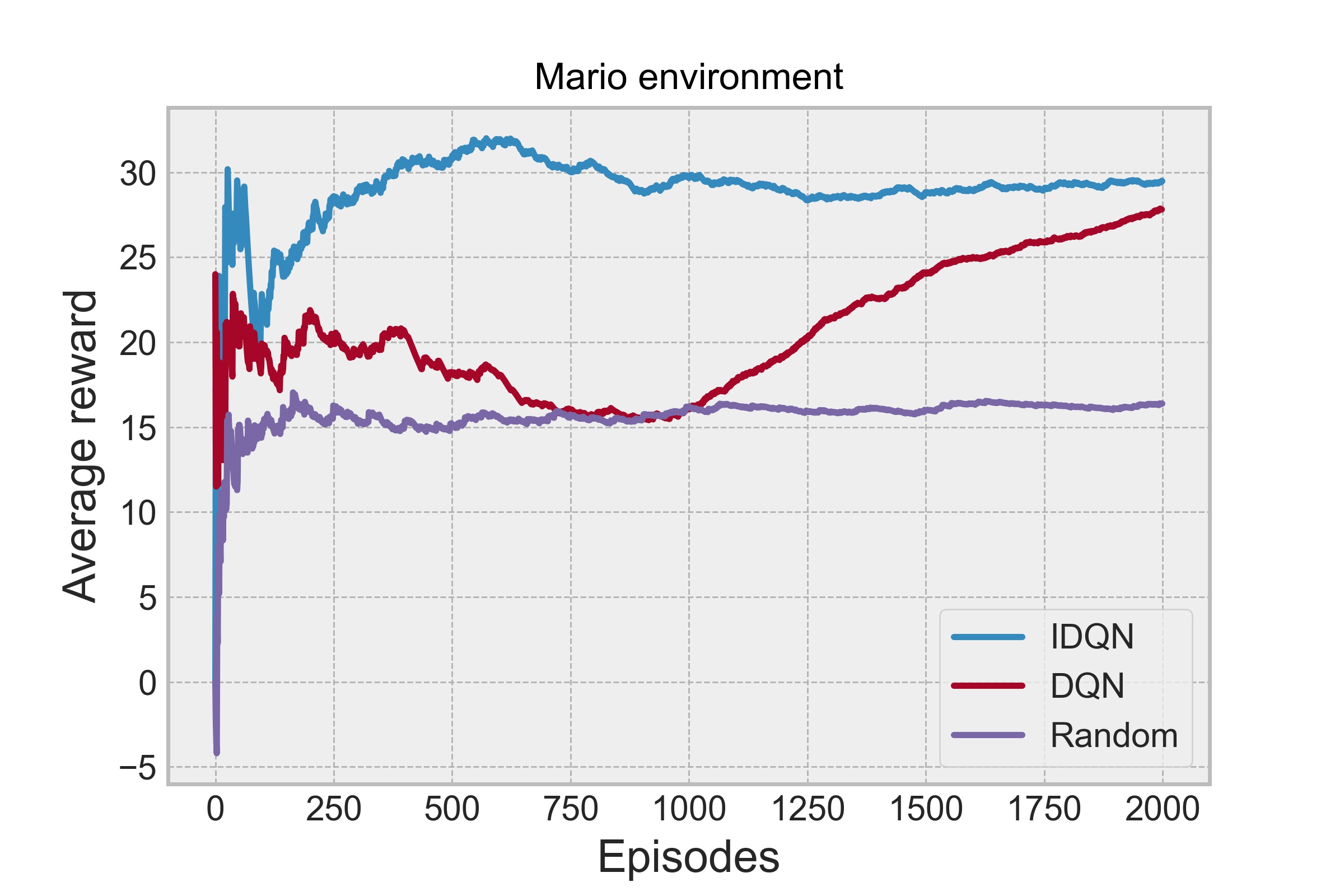}%
        \label{fig:results_a}}\\
        \subfloat[]{\includegraphics[width=1.5in]{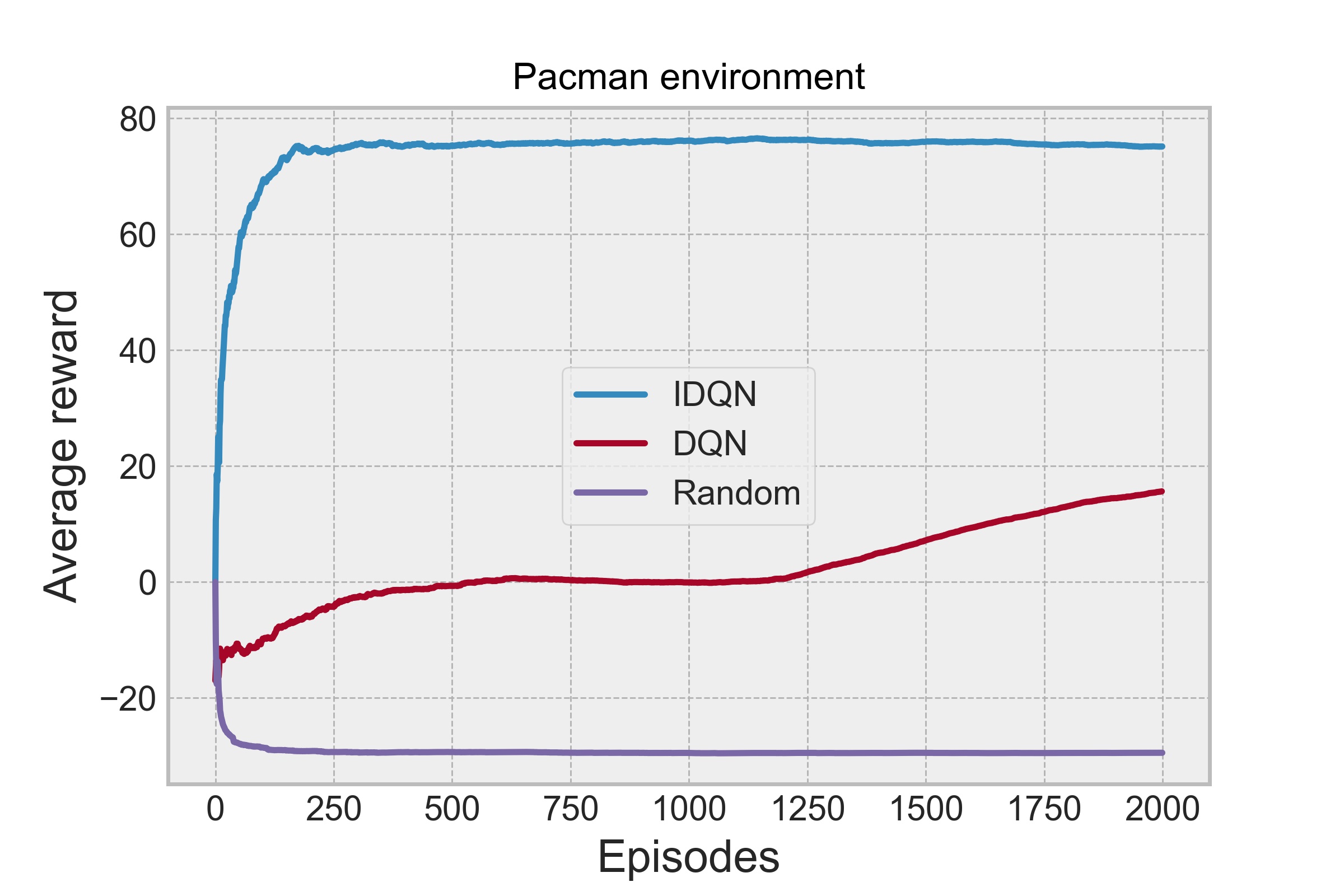}%
        \label{fig:results_b}}\\
        \subfloat[]{\includegraphics[width=1.5in]{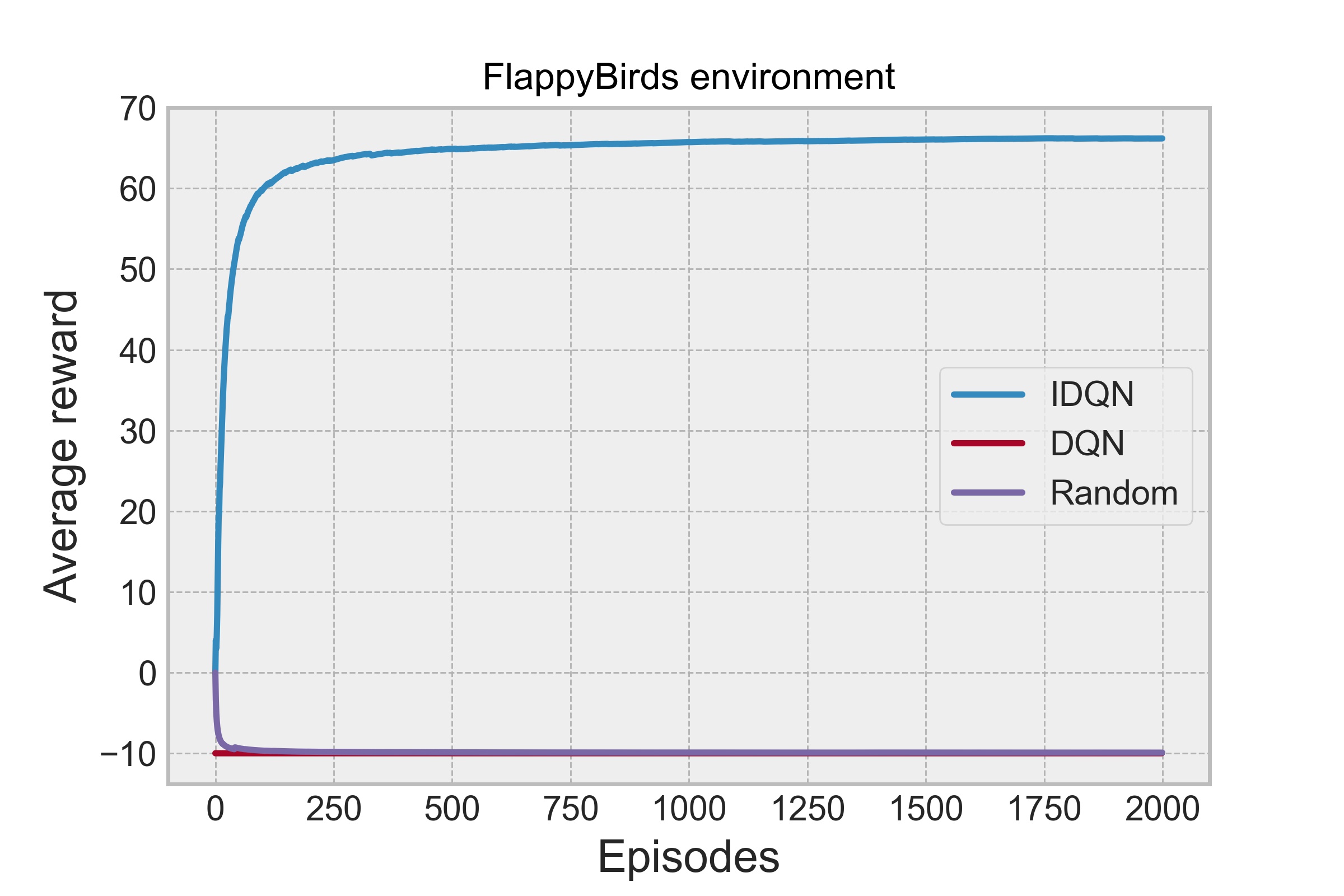}%
        \label{fig:results_c}}\\
        \subfloat[]{\includegraphics[width=1.5in]{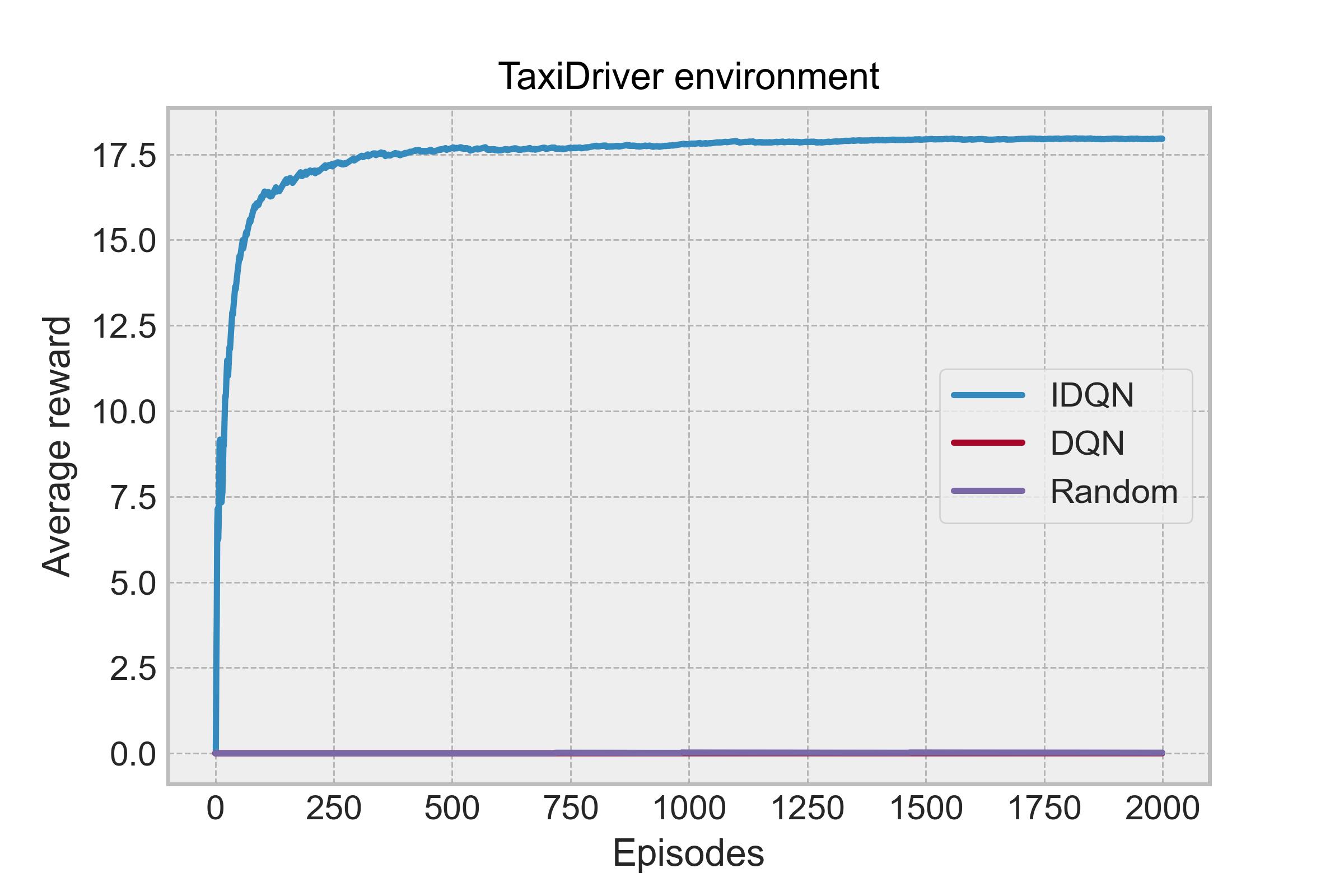}%
        \label{fig:results_d}}\\
        \subfloat[]{\includegraphics[width=1.5in,height=1.0in]{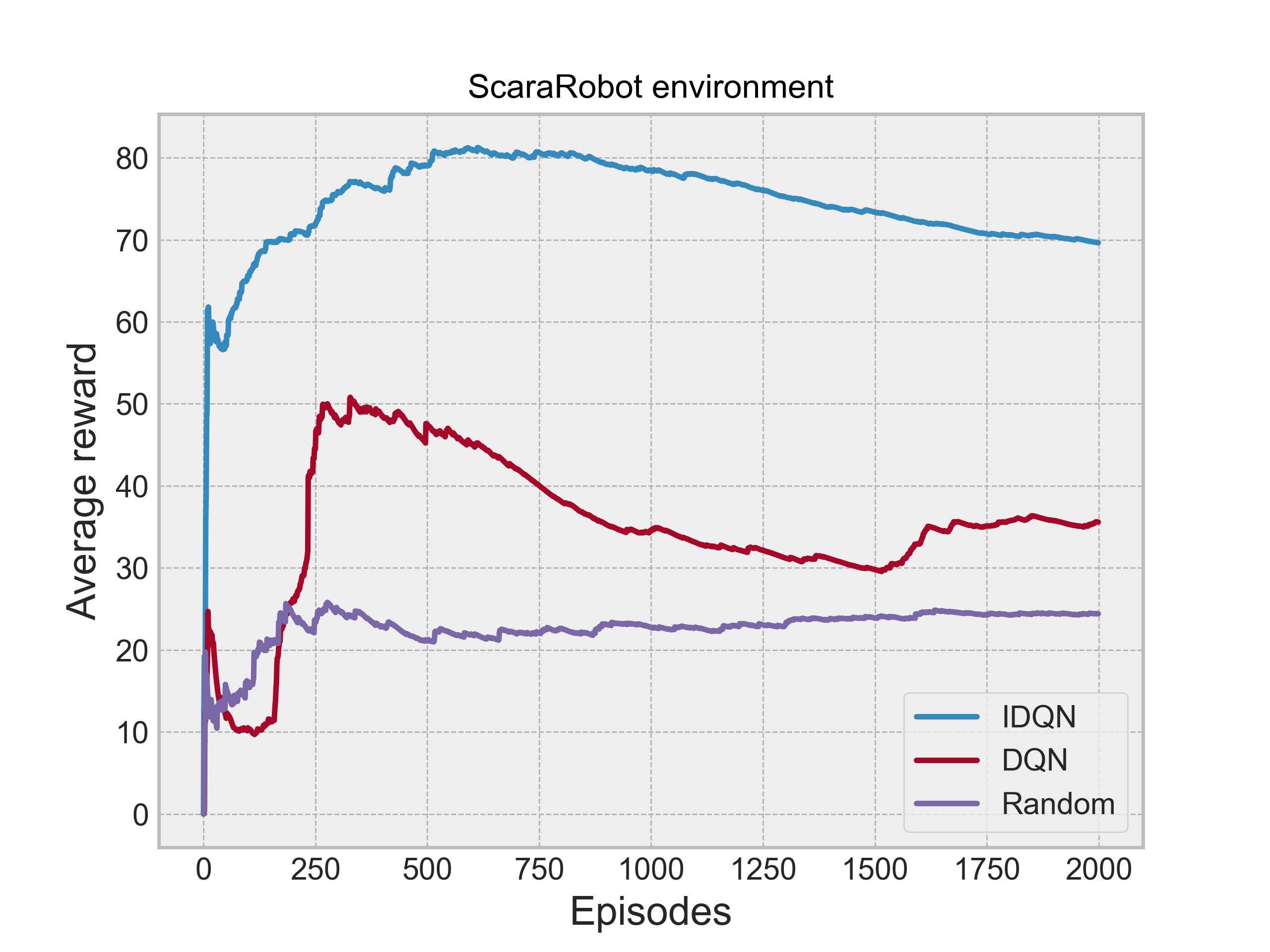}%
        \label{fig:results_e}}\\
        \end{tabular}
        &
        \begin{tabular}{c}
        \smallskip
        \subfloat[]{\includegraphics[width=1.5in]{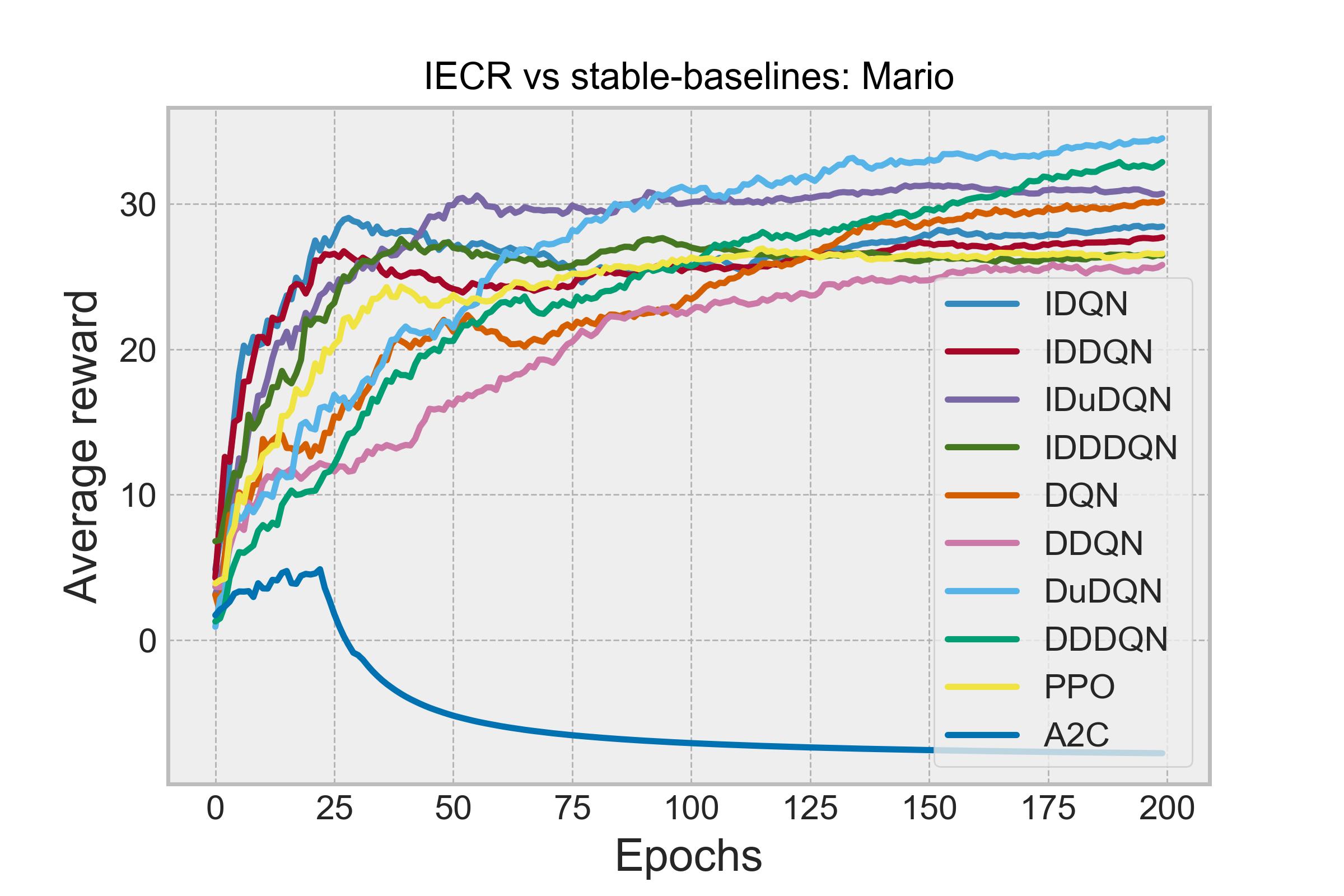}%
        \label{fig:results_f}}\\
        \subfloat[]{\includegraphics[width=1.5in]{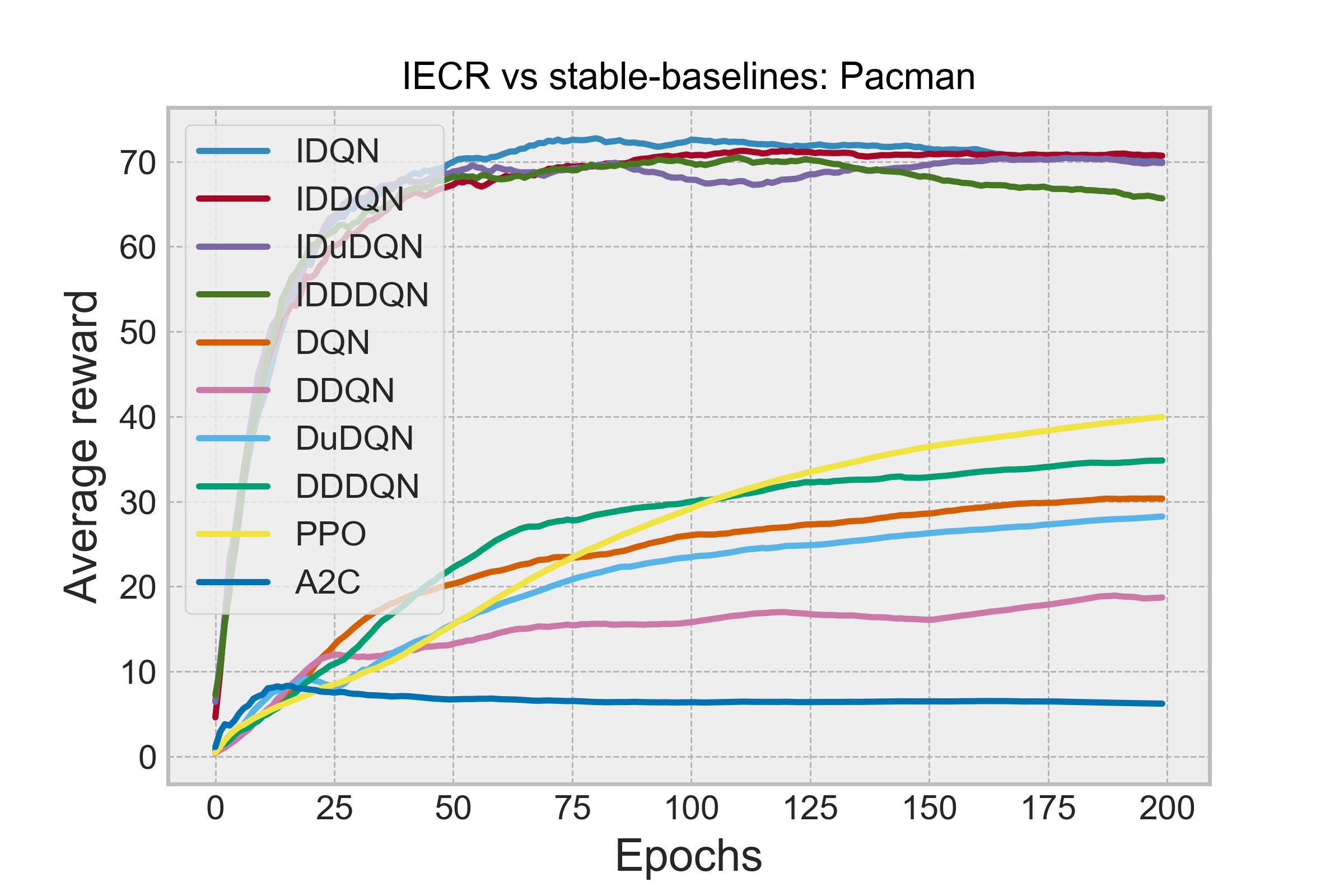}%
        \label{fig:results_g}}\\
        \subfloat[]{\includegraphics[width=1.5in]{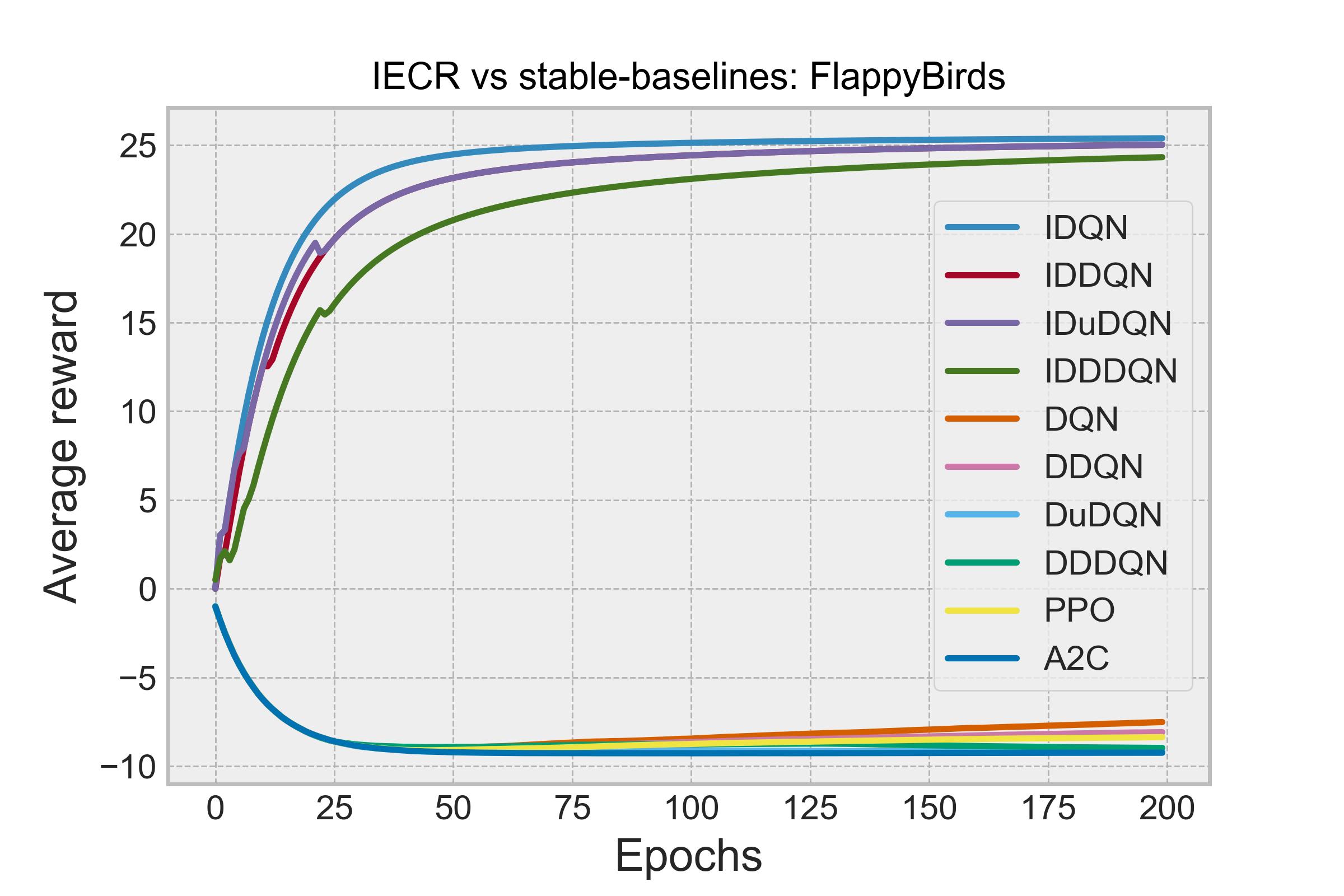}%
        \label{fig:results_h}}\\
        \subfloat[]{\includegraphics[width=1.5in,height=1.0in]{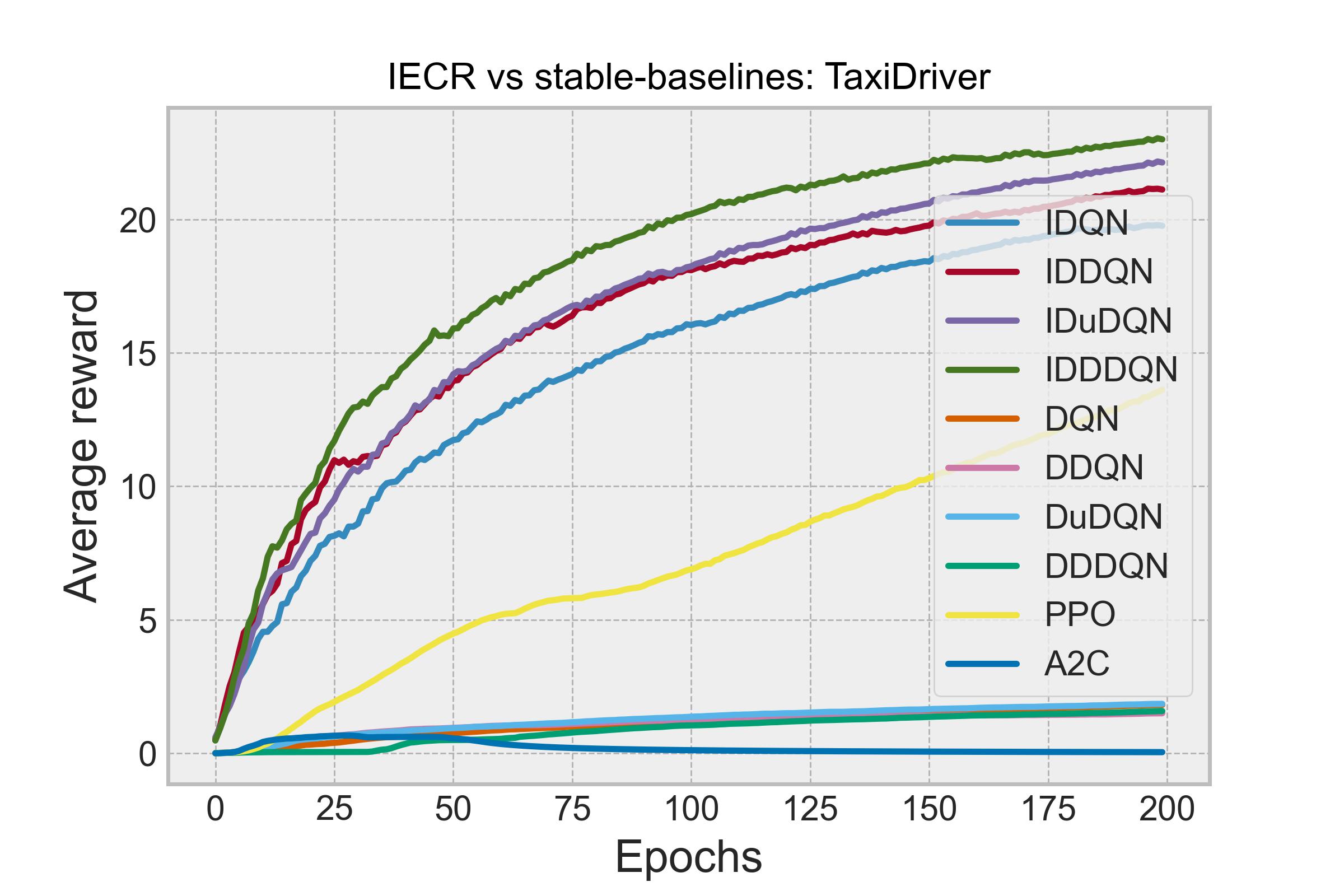}%
        \label{fig:results_i}}\\
        \subfloat[]{\includegraphics[width=1.5in]{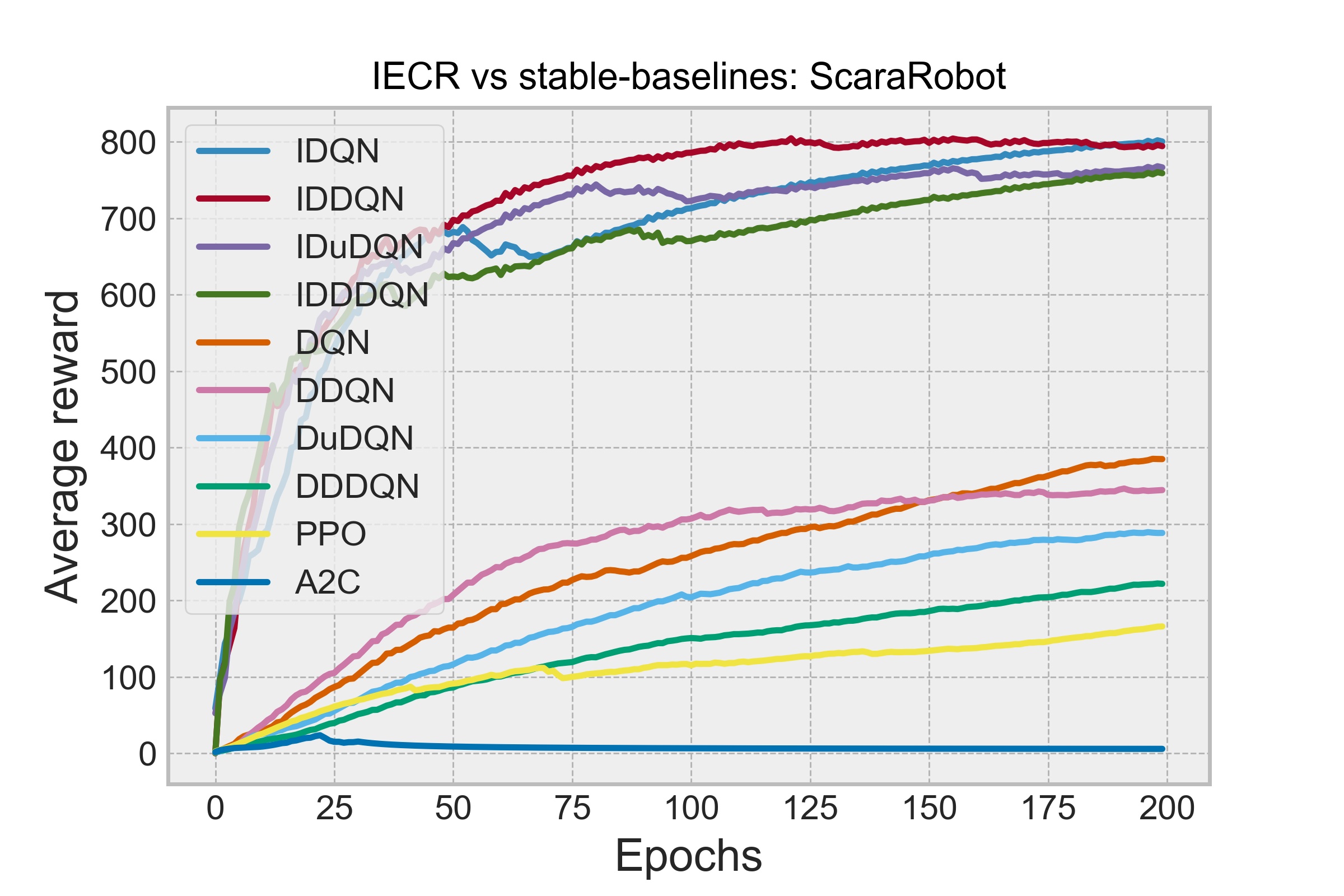}%
        \label{fig:results_j}}\\
        \end{tabular}
    \end{tabular}
\caption{The learning curves above are the results of the two stages of experiments. The results of the first stage of the experiment are shown in (a) Mario, (c) Pacman, (c) Flappybirds, (d) TaxiDriver, and (e) ScaraRobot. The progress is measured in every episode. The results of the second stage of the experiments are shown in (f), (g), (h), (i) and (j), where the progress is measured every epoch. Every epoch is equivalent to 400 training steps. In the second stage, the learning progress of the state-of-the-art algorithms was less chaotic. However, IECR variants still outperform the state-of-the-art approaches.}
\label{fig:results}
\end{figure}

From Fig.~\ref{fig:results_a} to Fig.~\ref{fig:results_e}, we show the learning curves of the first stage of the experiments for the five experimental environments: Mario, Pacman, FlappyBirds, TaxiDriver, and ScaraRobot. Both DQN and the random agents show similar performance within the Mario environment at the beginning of the training. The nature of the environment can explain this behavior. While Pacman is highly constrained, Mario can take several actions within its open environment. Meanwhile, when there is an enemy or an obstacle, our $\iota(s)$ function provides actions that prevent the agent from dying or getting stuck. Consequently, IDQN feeds the replay buffer with better-quality data while reaching further states during the game.

In the Pacman environment, the efforts of DQN are impractical because every episode ends prematurely. Likewise, when using a random policy, the performance is poor because, most of the time, the agent can only go in two directions. Consequently, the equal probability of taking those actions leads the agent to a poor exploration of the environment. IDQN performs better because our $\iota(s)$ function avoids impractical actions such as crashing against a wall or an enemy. This behavior may take more than 2,000 episodes for DQN to learn.

In the FlappyBirds environment, IDQN can easily outperform DQN because the stochastic nature of the $\epsilon$-greedy policy exploration with equal probability indirectly biases the agent's behavior. Therefore, the replay buffer fills with data related to crashes against the pipes, floor, or ceiling and rarely with passing through the pipes. Moreover, when randomly sampling a mini-batch from the replay buffer, the probability of picking a transition representing a positive reward when passing through the pipes is exceptionally low. Hence, the agent mainly trains with impractical data and rarely trains from transitions that represent the actual goal of the game. In contrast, IDQN can make decisions based on the obstacles around the bird. Thus, every episode lasts longer and fills the replay buffer with better-quality training data. 

In the TaxiDriver and ScaraRobot environments, DQN presents difficulty in learning the drop action. In these environments, it is very easy to collide against obstacles due to the stochastic nature of DQN. This behavior fills the buffer with useless data, and the episodes end prematurely. IDQN does not need to explore or learn when to pick or drop because that behavior is already encoded in the affordance function $\iota(s)$.

\subsection{Results of the second stage of the experiments}

For the second stage of the experiments, the learning curves from Fig.~\ref{fig:results_f} to Fig.~\ref{fig:results_h} show a comparison between the performance of the state-of-the-art algorithms DQN, DDQN, DuDQN, DDDQN, PPO, and A2C and the IECR variants IDQN, IDDQN, IDuDQN, and IDDDQN. We summarize the results of this stage in Table~\ref{table:results_b}. We applied the same number of training steps to compare the algorithms fairly. The agent was trained for 200 epochs, where every epoch is equivalent to 400 training steps. In the Mario environment, the DQN, DDQN, DuDQN, and DDDQN learning curves show more stable progress than in the first stage of the experiments. We can also observe better learning progress when $\iota(s)$ is involved. In Mario's environment, the gap between state-of-the-art approaches is smaller than in Pacman and FlappyBirds because this environment has plenty of open space where the agent can decide where to go. Moreover, the $\iota(s)$ only has an effect when enemies, pipes, or holes are nearby, and most of the time, all the actions are available. However, $\iota(s)$ assists in critical moments of decisions and allows the agent to explore further and fill the replay buffer with this information. Besides, PPO showed competitive learning progress, while A2C got stuck into a local minimum (see Fig.~\ref{fig:results_f}).

In the Pacman environment, $\iota(s)$ has a higher impact because environment obstacles such as walls and ghosts are always next to Pacman. While the state-of-the-art approaches use a stochastic policy for exploring the environment provoking Pacman to crash against the walls or to die by touching a ghost, $\iota(s)$ will avoid these states.

In the FlappyBirds environment, the state-of-the-art approaches had difficulties in exploring the environment and finding high-value states. On the contrary, $\iota(s)$ explores and supports good decisions based on what is surrounding the bird. Consequently, the state-of-the-art approaches rarely train from adequate states, whereas $\iota(s)$ guides the agent into better decisions through every episode.

In the TaxiDriver environment, the affordances function $\iota(s)$ assisted the agent exploration efficiently because, once the taxi reached the passenger position, the only possible action according to the set of rules is picking. Therefore, the state-of-the-art algorithms must visit the same state several times to learn that picking is the only possible action at that state. In Fig.~\ref{fig:results_i}, it can be noted that IECR variants learn faster due to the reason explained before.

The ScaraRobot environment behaves similarly to TaxiDriver. When the robotic arm reaches the picking or dropping positions, only one action is allowed (picking or dropping). However, the state-of-the-art approaches cannot produce enough high-quality data dependent on the action-selection process, which reverberates in their learning performance (Fig.~\ref{fig:results_f}).

\begin{table*}
\begin{center}
\caption{\label{table:results_b}Average reward obtained in the second stage of the experiments.}
\begin{tabular}{| p{0.1\textwidth} p{0.05\textwidth} p{0.05\textwidth} p{0.05\textwidth} p{0.05\textwidth} p{0.05\textwidth} p{0.05\textwidth} p{0.05\textwidth} p{0.05\textwidth} p{0.05\textwidth} p{0.05\textwidth} |}
 \hline
  & IDQN & IDDQN & IDuDQN & IDDDQN & DQN & DDQN & DuDQN & DDDQN & PPO & A2C \\
 \hline
  \hline
Mario & $26.24$ &$25.87$&$\textbf{30.24}$&$23.69$&$23.24$&$20.28$&$27.06$&$23.88$&$23.94$&$-5.24$\\

 Pacman & $45.77$ &$\textbf{45.03}$&$39.71$&$41.16$&$23.02$&$14.63$&$20.36$&$26.05$&$25.68$&$ 6.55$\\

 FlappyBirds & $\textbf{15.18}$&$15.18$&$14.16$&$13.49$&$-8.02$&$-8.26$&$-8.72$&$-8.49$&$-8.38$&$-8.83$\\

 TaxiDriver & $14.51$&$16.22$&$16.54$&$\textbf{18.1}$&$1.11$&$1.11$&$1.24$&$0.9$&$7.05$&$0.22$\\

 ScaraRobot & $690.35$&$\textbf{715.17}$&$679.27$&$647.82$&$239.26$&$258.12$&$182.97$&$133.04$&$108.48$&$7.83$\\
 \hline
\end{tabular}
\end{center}
\end{table*}

\subsection{Affordance loss impact}

\begin{figure}[htbp]
    \centering
            \includegraphics[width=3.5in]{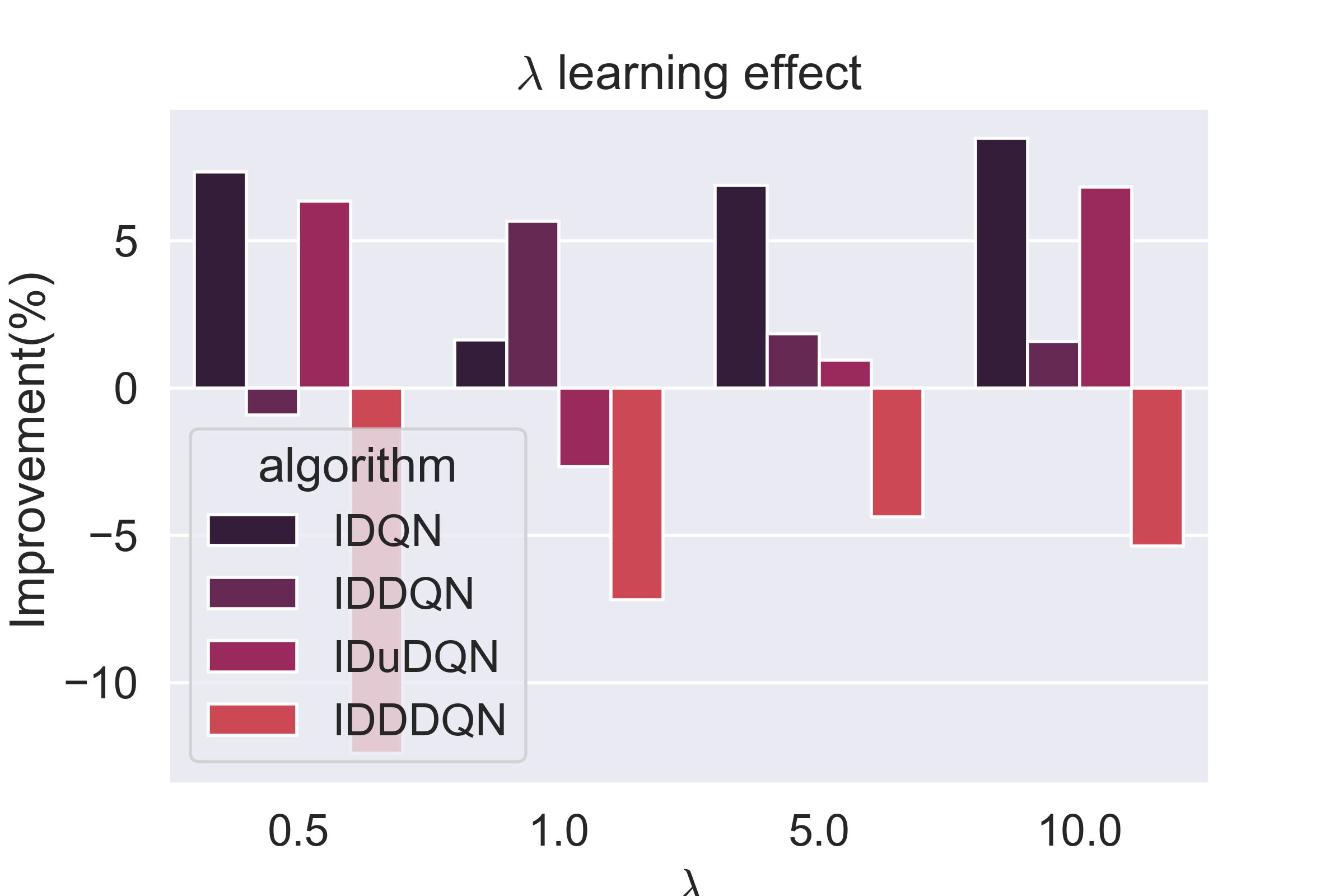}%
    \caption{The results above show the effect of varying the value of $\lambda$ in the algorithms.}
    \label{fig:lambda_effect}
\end{figure}

To measure the impact of the affordance loss, we calculated the percentage of improvement by taking 
$\lambda=0$ as a reference for all the environments. In Fig.~\ref{fig:lambda_effect}, we summarize the results showed in Table~\ref{table:results_lambda}. When $\lambda=0.5$, IDQN and IDuDQN had an improvement of $7.5\%$ and $6\%$, respectively. While IDDQN and IDDDQN showed a deterioration in the performance of $1\%$ and $14\%$. For $\lambda=1$, IDQN and IDDQN increased their performance by $1.5\%$ and $5.5\%$, whereas IDuDQN and IDDDQN performance reduced in $2.5\%$ and $7.5\%$, respectively. The experiment results show that $\lambda=5$ produced an increment in the performance of IDQN, IDDQN, and IDuDQN of $6.8\%$, $1.8\%$ and  $0.5\%$, while IDDDQN performance dropped in $4.5\%$. Setting $\lambda=10$ provoked that IDQN, IDDQN, and IDuDQN increased their performance by $8.9\%$, $1.7\%$ and $6.5\%$, respectively. While IDDDQN performance decreased $5.2\%$.

\begin{table}
\begin{center}
\caption{\label{table:results_lambda} Affordance loss impact (average reward).}
\begin{tabular}{ p{0.1\textwidth} p{0.05\textwidth} p{0.05\textwidth} p{0.05\textwidth} p{0.05\textwidth} |}
\hline
$\lambda=0$& IDQN & IDDQN & IDuDQN & IDDDQN  \\
\hline
\hline
Mario & $24.91$ &$25.01$&$26.75$&$\textbf{36.36}$\\

Pacman & $41.19$ &$45.92$&$\textbf{48.95}$&$41.02$\\

FlappyBirds & $14.93$&$\textbf{15.18}$&$13.38$&$14.41$\\

TaxiDriver & $14.77$&$15.29$&$15.51$&$\textbf{16.86}$\\

ScaraRobot & $680.65$&$641.92$&$\textbf{731.89}$&$697.07$\\
\hline\\

\hline
  $\lambda=0.5$& IDQN & IDDQN & IDuDQN & IDDDQN  \\
 \hline
  \hline
Mario & $25.65$ &$\textbf{26.81}$&$23.46$&$23.81$\\

 Pacman & $42.78$ &$\textbf{47.58}$&$45.86$&$42.95$\\

 FlappyBirds & $14.9$&$\textbf{15.18}$&$13.76$&$13.92$\\

 TaxiDriver & $14.97$&$15.46$&$\textbf{15.91}$&$15.8$\\

 ScaraRobot & $810.28$&$595.5$&$\textbf{966.78}$&$573.0$\\
  \hline\\

\hline 
  $\lambda=1$& IDQN & IDDQN & IDuDQN & IDDDQN  \\
\hline 
\hline
Mario & $26.24$ &$25.87$&$\textbf{30.24}$&$23.69$\\

 Pacman & $\textbf{45.77}$ &$45.03$&$39.71$&$41.16$\\

 FlappyBirds & $\textbf{15.18}$&$15.18$&$14.16$&$13.49$\\

 TaxiDriver & $14.51$&$16.22$&$16.54$&$\textbf{18.1}$\\

 ScaraRobot & $690.35$&$\textbf{715.17}$&$679.27$&$647.82$\\
 \hline\\

\hline
  $\lambda=5$& IDQN & IDDQN & IDuDQN & IDDDQN  \\
 \hline
  \hline
Mario & $\textbf{26.42}$ &$24.54$&$24.79$&$25.75$\\

 Pacman & $41.96$ &$\textbf{50.42}$&$45.47$&$50.09$\\

 FlappyBirds & $\textbf{15.18}$&$15.18$&$13.49$&$13.33$\\

 TaxiDriver & $14.94$&$\textbf{16.64}$&$14.5$&$15.27$\\

 ScaraRobot & $841.11$&$593.11$&$\textbf{914.08}$&$711.51$\\
  \hline\\
\hline
  $\lambda=10$& IDQN & IDDQN & IDuDQN & IDDDQN  \\
 \hline
  \hline
Mario & $25.14$ &$26.12$&$24.31$&$\textbf{29.34}$\\

 Pacman & $44.43$ &$\textbf{48.63}$&$44.73$&$41.87$\\

 FlappyBirds & $\textbf{15.16}$&$15.12$&$13.75$&$13.29$\\

 TaxiDriver & $16.83$&$16.1$&$18.3$&$\textbf{18.55}$\\

 ScaraRobot & $802.81$&$591.69$&$\textbf{959.26}$&$614.2$\\
 \hline
\end{tabular}
\end{center}
\end{table}

\section{Discussion}
\label{sec:discussion}
This paper introduced the IECR framework, aiming to accelerate the agent's learning process by utilizing available contextual information in the environment. This approach is first fed from a manually defined set of rules, much like what a human does. Then it uses the rules to generate an affordance function that seeks to reduce the exploration space and optimize the decision-making process.

In real-world tasks, humans understand the rules, so repeating the same mistake millions of times is unnecessary. IECR is a successful framework that uses this human capability to benefit from context and use it to make intelligent decisions. The fact that IECR variants outperform all these state-of-the-art approaches makes it clear how vital context is during the learning process of an agent.

In the FlappyBirds environment, the IECR variants IDQN, IDDQN, IDuDQN, and IDDDQN outperform state-of-the-art approaches. There may be better approaches than DQN, DDQN, DuDQN, or DDDQN. For example, a Q-table and QL could manage to solve this environment quickly. Since FlappyBirds is a highly repetitive environment, the number of states can be calculated. Another solution may be to use two replay buffers: one buffer for poor and average transitions and a second buffer exclusive for good transitions. It would be necessary to bias the sampling process of the data to collect good transitions so that the agent can learn and find a solution for the environment. Another solution may be to design a reward function that pushes the bird to the middle of the pipes. However, IDQN does not require such modifications, and a set of rules is enough to solve the environment. 

The results in Table~\ref{table:results_lambda} show that $\lambda$ and the two loss functions based on the affordance function ($J(Q)_{simple}$ and $J(Q)_{double}$) have a higher impact in IDQN, IDDQN, and IDuDQN than in IDDDQN. We believe that the reason for this is that the $J(Q)_{double}$ loss function is related to the Q-values stream of the target neural network. Hence, the weights of the main and target neural networks may provoke the possible actions in the advantage stream to differ from the ones in the Q-values stream.

Simply carrying out a stochastic exploration of the environment does not secure a good learning process for the agent. Therefore, the satisfactory results of IECR introduced in this paper can be explained by the quality of the data produced with the assistance of the $\iota(s)$ function. IECR  has the potential to be implemented not only in games but in other domains because it is compatible with discrete environments.  

\section{Conclusion and future work}\label{sec:conclusion}

The framework developed in this paper demonstrates a seamless integration of contextual data in deep RL. We have demonstrated that our framework and the resulting algorithms, namely IDQN, IDDQN, IDuDQN, and IDDDQN, significantly enhance the agent's performance by converging in the experimental environments within approximately 40,000 training steps. In comparison, all state-of-the-art approaches in this study exhibited inferior results when compared to IECR variants. By incorporating a manually defined set of rules and the concept of CKFs, we obtain an affordance function, enabling the agent to benefit from higher-quality data compared to the data obtained from a pure stochastic exploration. Our contributions include an intuitive framework that incorporates contextual information to improve RL, as well as the introduction of four novel algorithms based on IECR.

The results from the first stage of experiments demonstrate the capacity of IDQN to learn from CKFs representation. In the second stage, our results indicate that IECR variants consistently outperform state-of-the-art algorithms when utilizing the same number of training steps (40,000) as a reference. IECR provides a solution for challenging states where, instead of learning after millions of interactions, the agent computes $\iota(s)$ and avoids becoming stuck before proceeding to explore the environment.

This paper has explored the integration of context in RL and has opened new avenues for challenges in the field. The current limitations of this work pertain to continuous environments, where the state representation may pose difficulties in finding an affordance function. In future work, we plan to investigate the application of this framework in continuous and three-dimensional environments, with a particular focus on robotics. We hypothesize that our affordance function, $\iota(s)$, which represents the probability of taking an action based on the context, could also accelerate the learning process of agents in these settings. Moreover, with the semantic representation of the environment, there is potential to develop an agent capable of learning, explaining its actions, and understanding the environment.

\section*{Acknowledgment}

This project was partially supported by  Consejo Nacional de Humaninades Ciencias y Tecnologías (CONAHCyT) and Cardiff University. 

{\appendices
}

\bibliographystyle{IEEEtran}
\bibliography{bibliography}

\begin{IEEEbiography}[{\includegraphics[width=1in,height=1.25in,clip,keepaspectratio]{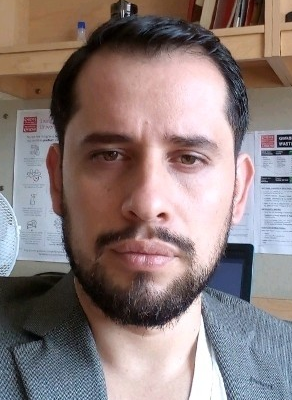}}]{Francisco Munguia-Galeano} received the B.S. degree in robotics engineering and his master's degree (Hons.) in computing technology from Instituto Politécnico Nacional, Mexico, in 2015 and 2018, respectively. He is currently pursuing the Ph.D. degree in engineering at Cardiff University, United Kingdom. Prior to joining Cardiff University as a Research Assistant in 2021, he gained experience working as an Automation Engineer and, more recently, as a software developer. His research interests encompass reinforcement learning and robotics.
\end{IEEEbiography}

\begin{IEEEbiography}[{\includegraphics[width=1in,height=1.25in,clip,keepaspectratio]{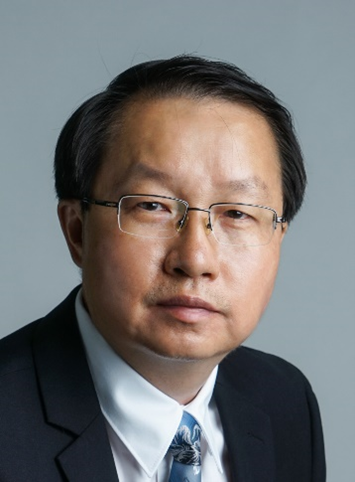}}]{Ah Hwee Tan}(SM’04) received the B.Sc. (Hons.) and M.Sc. degrees in computer and information science from the National University of Singapore, Singapore, and the Ph.D. degree in cognitive and neural systems from Boston University, Boston, MA, USA. He is currently Professor of Computer Science and Associate Dean of Research at the School of Computing and Information Systems, Singapore Management University (SMU). His current research interests include cognitive and neural systems, brain-inspired intelligent agents, machine learning, knowledge discovery, and text mining. Dr Tan is an Associate Editor of IEEE TRANSACTIONS ON NEURAL NETWORKS AND LEARNING SYSTEMS, Vice Chair of IEEE CIS Task Force on Towards Human-Like Intelligence, and a member of IEEE CIS Neural Networks Technical Committee (NNTC).
\end{IEEEbiography}

\begin{IEEEbiography}[{\includegraphics[width=1in,height=1.25in,clip,keepaspectratio]{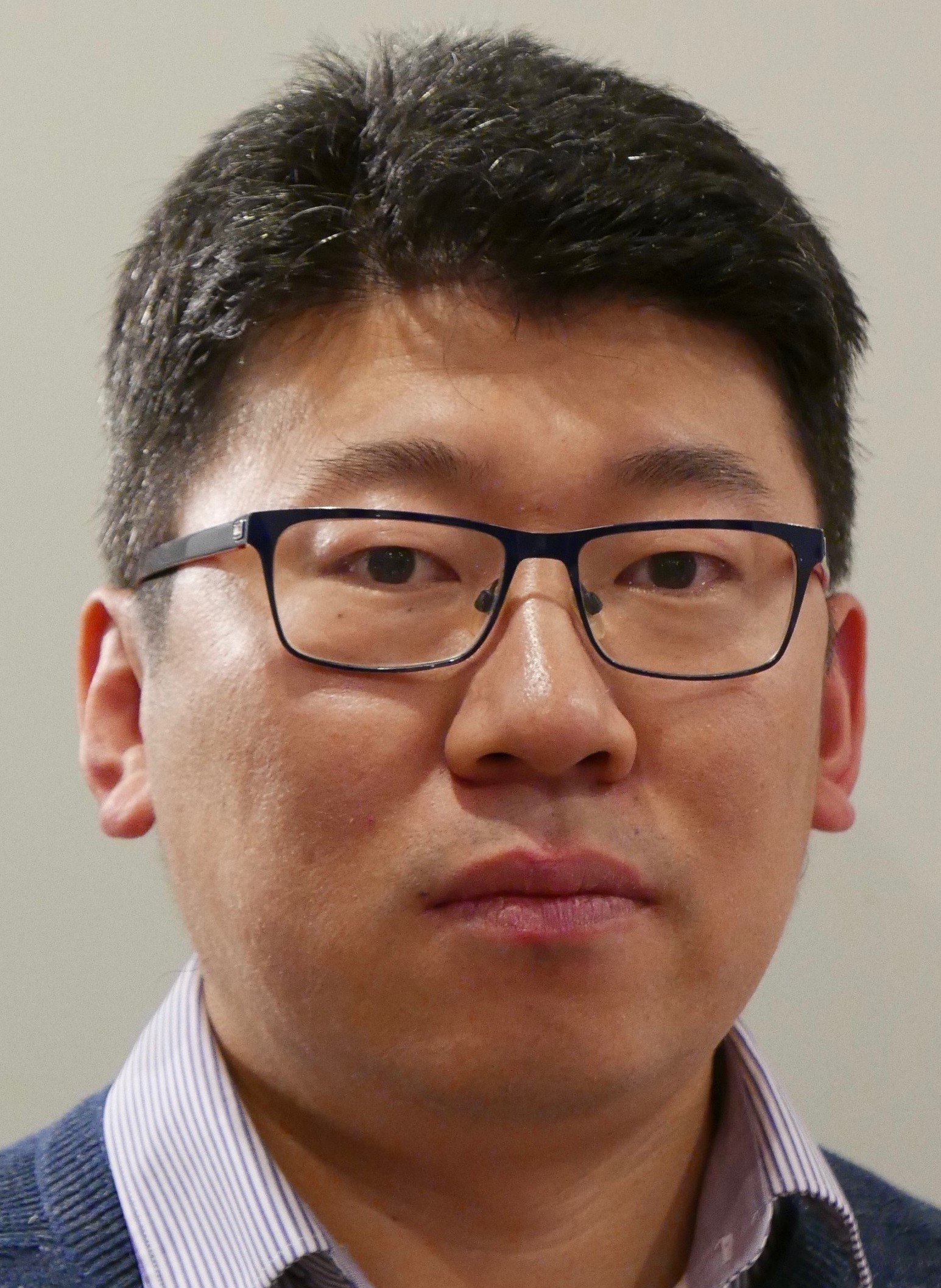}}]{Ze Ji}(Member, IEEE) received a B.Eng. degree from Jilin University, Changchun, China, in 2001, M.Sc. degree from the University of Birmingham, Birmingham, U.K., in 2003, and Ph.D. degree from Cardiff University, Cardiff, U.K., in 2007. He is a senior lecturer (associate professor) with the School of Engineering, Cardiff University, U.K. He is also the recipient of the Royal Academy of Engineering Industrial Fellow. Prior to his current position, he was working in industry (Dyson, Lenovo, etc) on autonomous robotics. His research interests are cross-disciplinary, including autonomous robot navigation, robot manipulation, robot learning, computer vision, simultaneous localization and mapping (SLAM), acoustic localization, and tactile sensing.
\end{IEEEbiography}
\vspace{11pt}

\end{document}